%% file: PhongSurface_eccv2020.tex
\documentclass[runningheads]{llncs}

\usepackage{graphicx}
\usepackage{comment}
\usepackage{amsmath,amssymb} 
\usepackage{xcolor}

\usepackage{cite}
\usepackage[export]{adjustbox}

\input{preamble}

\begin{document}
\pagestyle{headings}
\mainmatter
\def\ECCVSubNumber{2597}  

\title{The Phong Surface: Efficient 3D Model Fitting using Lifted Optimization}


\titlerunning{Phong surface model}
\author{Jingjing Shen \and
Thomas J. Cashman \and
Qi Ye \and
Tim Hutton \and
Toby Sharp \and
Federica Bogo \and
Andrew Fitzgibbon \and
Jamie Shotton}
\authorrunning{J. Shen et al.}
\institute{Microsoft Mixed Reality \& AI Labs, Cambridge, UK \\
\email{\{jinshen, tcashman, yeqi, tihutt, tsharp, febogo, awf, jamiesho\}@microsoft.com}}
\maketitle

\begin{abstract}

Realtime perceptual and interaction capabilities in mixed reality require a range of 3D tracking problems to be solved at low latency on resource-constrained hardware such as head-mounted devices.
Indeed, for devices such as HoloLens 2 where the CPU and GPU are left available for applications, multiple tracking subsystems are required to run on a continuous, real-time basis while sharing a single Digital Signal Processor.
To solve model-fitting problems for HoloLens 2 hand tracking, where the computational budget is approximately 100 times smaller than an iPhone 7, we introduce a new surface model: the `Phong surface'.
Using ideas from computer graphics, the Phong surface describes the same 3D shape as a triangulated mesh model, but with continuous surface normals which enable the use of lifting-based optimization, providing significant efficiency gains over ICP-based methods.
We show that Phong surfaces retain the convergence benefits of smoother surface models, while triangle meshes do not.

\end{abstract}

\input{Introduction}
\input{Method}

\input{Experiments}

\input{Conclusion}

\clearpage
%
%
\bibliographystyle{splncs04}
\bibliography{egbib}

\input{PhongSurfaceSupplementary}

\end{document}

%% file: preamble.tex
\usepackage{times}
\usepackage{epsfig}
\usepackage{graphicx}
\usepackage{amsmath}
\usepackage{amssymb}

\usepackage{url}

\usepackage{zlmtt}
\usepackage{upgreek} 
\usepackage{graphicx}
\usepackage{comment}
\usepackage{algorithm}
\usepackage{microtype}
\usepackage[noend]{algpseudocode}
\usepackage{amssymb}
\usepackage{booktabs}
\usepackage{xspace}
\usepackage{mathtools}
\usepackage{subcaption}

\usepackage[pagebackref=true,breaklinks=true,letterpaper=true,colorlinks,bookmarks=false]{hyperref}

\usepackage{tikz}
\usetikzlibrary{pgfplots.groupplots}

\usepackage{pgfplots}

\tikzstyle{ldotted}=                  [dash pattern=on 3pt off 2pt]

\definecolor{chira1}{HTML}{A02020} 
\definecolor{chira1b}{HTML}{ff4500} 
\definecolor{chira1c}{HTML}{ff6666} 

\definecolor{chira2}{HTML}{20A020} 
\definecolor{chira2c}{HTML}{006400} 
\definecolor{chira2b}{HTML}{66ee55} 

\definecolor{chira3}{HTML}{A0A020} 
\definecolor{chira3b}{HTML}{daa520} 
\definecolor{chira3d}{HTML}{000000} 
\definecolor{chira3c}{HTML}{a0a0a0} 

\definecolor{chira4}{HTML}{2020A0}
\definecolor{chira5}{HTML}{A020A0}
\definecolor{chira6}{HTML}{20A0A0}
\definecolor{chira7}{HTML}{A0A0A0}

\definecolor{chira8}{HTML}{ffffff}
\definecolor{chira9}{HTML}{ffffff}

\pgfplotscreateplotcyclelist{chira}{%
	{chira1},
	{chira2},
	{chira3},
	{chira4},
	{chira5},
	{chira6},
	{chira7},
	{chira8},
	{chira1b},
{chira2b},
{chira3b},	
	{chira1c},
{chira2c},
{chira3c},
	{chira1b,dotted},
	{chira2b,dotted},
	{chira3b,dotted},
	{chira1,scatter},
	{chira2,scatter},
	{scatter, only marks}
}
\pgfplotsset{
	compat=1.12,
	every axis/.append style={
		grid=major,
		axis lines=center,
		scale only axis,
		cycle list name=chira,
		font=\scriptsize,
		xticklabel style={/pgf/number format/fixed, font=\scriptsize,},
		yticklabel style={/pgf/number format/fixed, font=\scriptsize,},
		scaled ticks=false,
		every axis x label/.style={
			at={(ticklabel cs:0.5)},
			anchor=near ticklabel,
			font=\scriptsize,
		},
		every axis y label/.style={
			at={(ticklabel cs:0.5)},
			rotate=90,
			anchor=near ticklabel,
			font=\scriptsize,
		},
		legend image code/.code={
			\draw[mark repeat=2,mark phase=2,very thick,##1] plot coordinates { (0cm,0cm) (0.15cm,0cm) (0.3cm,0cm) };%
		}
	},
	every axis legend/.append style={
		legend pos=south east,
		legend cell align=left,
		font=\tiny,
		draw=none,
	},
	every axis plot/.append style={
		thick
	}
}

\usetikzlibrary{arrows.meta}
\newlength{\plotheight}
\newcommand{\be}{\begin{equation}}
\newcommand{\ee}{\end{equation}}

\newcommand{\averagemetriclabel}[1]{\% of dataset with avg. \\ joint err. $< #1$}
\newcommand{\maxmetriclabel}[1]{\% of dataset with max. \\ joint err. $< #1$}

\newcommand{\etal}{et al.\xspace}

\DeclareMathOperator*{\argmin}{argmin}

\newlength{\yaxiswidth}

%% file: Introduction.tex

\newcommand{\strategy}[1]{#1}

\section{Introduction}

As computer vision systems are increasingly deployed on wearable or mobile computing platforms, they are required to operate with low power and limited computational resources.
In this context, the problem of pose estimation (as applied, for example, to tracking hands~\cite{Taylor2017, Mueller2019}, human bodies~\cite{Bogo2016, Xiang2019, Pavlakos2019}, or faces~\cite{Li2017}) is often tackled with a hybrid architecture that combines discriminative machine-learnt models with generative model fitting to explain the observed data~\cite{Taylor2016, Tkach2016, Taylor2017,  Mueller2019,  MagicLeap2019}.
For the purposes of this paper, we define `model fitting' as the registration of a 3D surface model to a point set observation.
Model fitters can benefit from powerful priors learned from data, and recent work even shows the benefits of including model fitting in the training loop~\cite{Nikos2019, Wan2019}.
An optimizer with fast convergence is critical for building real-time systems that operate with low compute. 
However, the {\em correspondences} between the observed data and the model are often unknown, and need to be discovered in the course of the optimization.

Two main optimization alternatives have been proposed for solving this problem.
\textbf{Iterative Closest Point (ICP)} algorithms~\cite{Besl1992, Hirshberg2012, Qian2014, Tagliasacchi2015} solve for model pose via `block coordinate descent': first finding closest points on the model surface, and then fixing those correspondences while solving for model pose alone.

The alternative approach is \textbf{`lifted' optimization}: to solve for correspondences and model pose \emph{simultaneously}, using a lifted objective function that explicitly parametrizes the unknown correspondences.
Taylor \etal~\cite{Taylor2014, Taylor2016} demonstrate hand tracking systems using this approach, and claim that the smoothness of the model surface is an important prerequisite for a smooth energy landscape that allows lifted optimization to converge efficiently.
However, the complex surface representation they propose comes at a cost, entailing as much as 58\% of the per-iteration model-fitting time~\cite[supplementary material]{Taylor2016}.

\begin{figure}[t]
	\centering
	\begin{subfigure}{0.2\linewidth}
		\centering
		\includegraphics[width=0.9\linewidth]{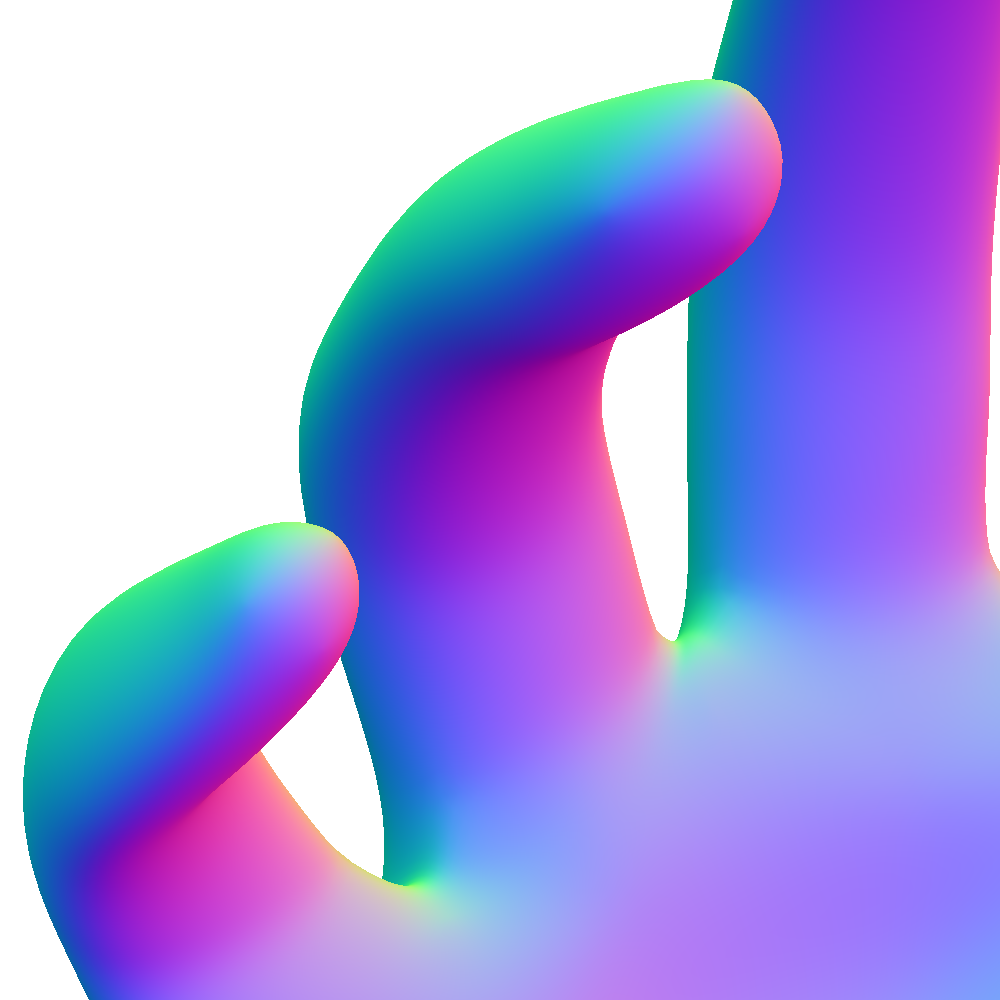}
		\caption{Subdiv. surface}
		\label{subfig:surfaces-loop}
	\end{subfigure}
	\begin{subfigure}{0.2\linewidth}
		\centering
		\includegraphics[width=0.9\linewidth]{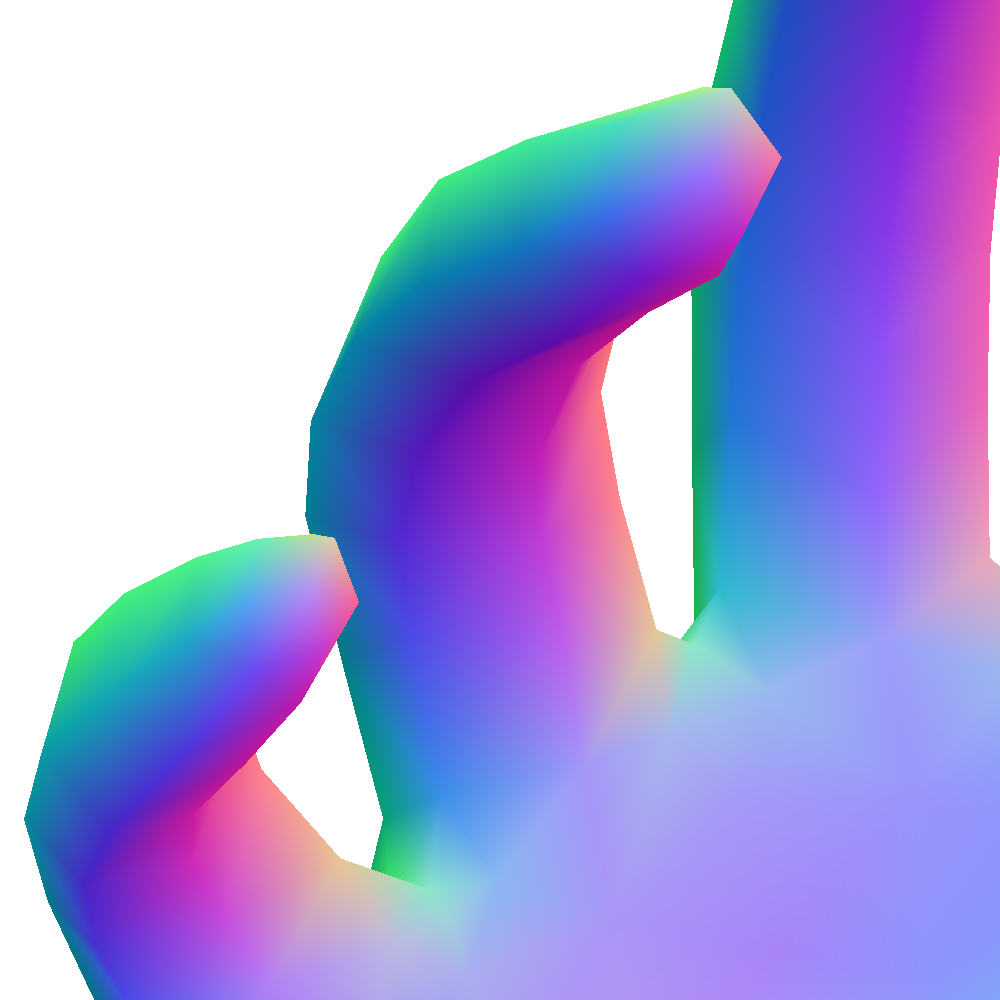}
		\caption{Phong surface}
		\label{subfig:surfaces-phong}
	\end{subfigure}
	\begin{subfigure}{0.2\linewidth}
		\centering
		\includegraphics[width=0.9\linewidth]{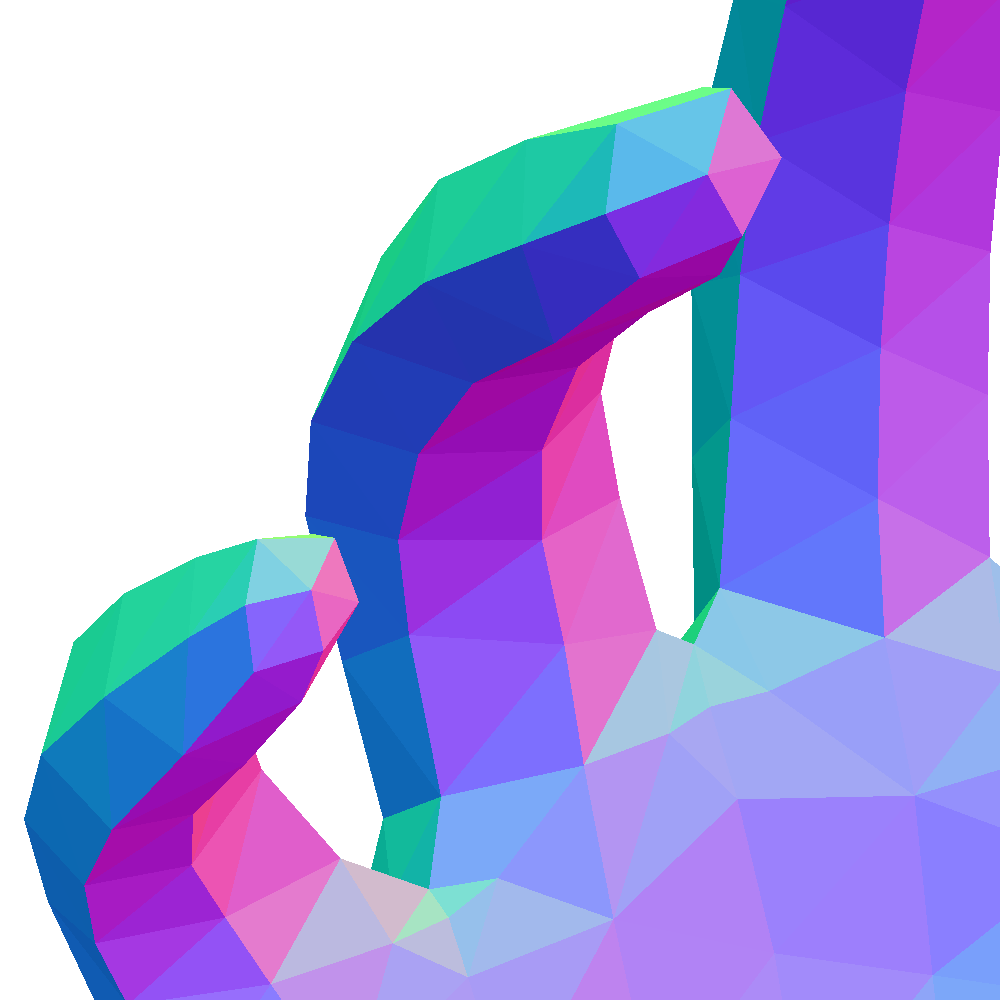}
		\caption{Triangular mesh}
		\label{subfig:surfaces-planar}
	\end{subfigure}
\begin{subfigure}{0.38\linewidth}
\begin{tabular}{lcc}\toprule
	Surface type
	&  eval.\
	&  with $\partial / \partial u$ \\
	\midrule
	Subdiv. surface & 0.329s  & 1.241s \\
	Phong surface  & 0.049s  & 0.279s \\
	Triangular mesh  & 0.047s & 0.196s \\
	\bottomrule
\end{tabular}
\caption{Timings on $10^6$ evaluations}
\label{subfig:surfaces-timing}
\end{subfigure}

\caption{A hand model represented by a Loop subdivision surface~\cite{Loop1987} (a), a Phong surface (b) and a triangle mesh (c), with surface normals visualized by mapping x, y, z coordinates to red, green and blue components respectively. (d) shows timings in seconds on PC: \emph{eval.} refers to evaluation of $10^6$ surface point positions and normals, and \emph{with $\partial / \partial u$} includes the cost of derivative calculations w.r.t. surface coordinate as well.}
\label{fig:surfaces}
\end{figure}

In this paper, we show that much simpler surface representations are sufficient
, thus making it cheaper for vision systems to access the convergence benefits of lifted optimization.
In particular, since it is often beneficial to include surface orientation properties in the objective function, one might assume that a normal field with continuous first derivatives (second-order surface smoothness) is a necessary minimum for a gradient-based optimizer.
This implies that a simple triangle mesh cannot suffice and we show this is true: a simple triangle mesh is insufficient, but it {\em is} sufficient to pair a triangle mesh with a normal field that is simply linearly interpolated over each triangle.
We call the resulting representation a `Phong surface', after the Phong shading~\cite{Phong1975} technique in computer graphics, which evaluates the lighting equation using similarly interpolated normals.
Fig.~\ref{fig:surfaces} shows the normal field evaluated inside the optimizer for different surface representations, illustrating that the Phong surface (b) leads to surface evaluations that are a close approximation to evaluations on the smooth subdivision surface (a).
Informally, our contribution to 3D model fitting is a surface representation which is designed to be ``as simple as possible, but no simpler".

We evaluate Phong surfaces in comparison to two other model representations; the smooth subdivision surfaces used by prior work~\cite{Cashman2013, Taylor2014, Taylor2016} and simple triangle meshes with a piecewise-constant normal field. 
Our experiments in rigid pose alignment and in the example application of hand tracking compare these alternative models in the context of different optimizers, and confirm earlier results that lifted optimization converges faster and with a wider basin of convergence than ICP.

We have successfully applied lifted optimization with Phong surfaces (Section 3.1) to implement a fully articulated hand tracker on HoloLens~2~\cite{HoloLens2019}, a self-contained head-mounted holographic computer.
{The tracker needs to run on an embedded Digital Signal Processor (DSP) with a compute budget of 4 GFLOPS, which is 100 times smaller than an iPhone 7, or 1000 times smaller than a high-end laptop.}  
This application required real-time continuous tracking under tight thermal and power constraints, motivating us to find the simplest possible computation that would allow the optimizer to converge to an accurate hand pose estimate.
The Phong surface representation is one of the key innovations that makes this hand tracker run in realtime on the HoloLens~2.

In summary, our contributions are that we
\begin{itemize}
\item introduce Phong surfaces by transferring the concept of Phong shading from computer graphics to model fitting in computer vision;
\item show that Phong surfaces combine the convergence benefits of lifted methods with the computational cost of planar mesh models;
\item demonstrate that fitting Phong surface models allows us to implement realtime hand tracking under a computational budget of 4 GFLOPS.
\end{itemize}

\subsection{Related work}

ICP has a long history in surface reconstruction and point set registration, starting from its first descriptions by Besl~and McKay~\cite{Besl1992} and Chen~and Medioni~\cite{Chen1992}.
The simplicity of ICP means that it is easy to implement and was broadly adopted, spawning a host of variations on the two fundamental steps of closest point finding and error minimization, as described by Rusinkiewicz~and Levoy~\cite{Rusinkiewicz2001}.
Neugebauer~\cite{neugebauer1997} and Fitzgibbon~\cite{fitzgibbon2003} formulate ICP as an instance of a non-linear least squares problem, solved by general optimizers such as the Levenberg-Marquardt algorithm~\cite{marquardt1963}.
This opens up new possibilities for the objective function, such as allowing a robust kernel to be included directly in the objective as a way to smoothly handle outliers~\cite{fitzgibbon2003}.

Meanwhile, ICP was generalized to articulated models by Pellegrini \etal~\cite{Pellegrini2008}, and demonstrated in this form for fitting models of human bodies~\cite{Zheng2014} and hands~\cite{Tagliasacchi2015}.
Another key advantage of ICP is the broad range of geometric representations that can be fitted: any point set or surface that could support the chosen `closest point' query is included in a straightforward manner.
However, a disadvantage of the alternating coordinate descent is that ICP algorithms \emph{converge slowly}, reducing the objective only \emph{linearly} as optimization proceeds~\cite{Taylor2017}.
This is particularly apparent when a model needs to slide relative to a data set, as this requires that the model's pose is updated in harmony with data correspondences.
Point-to-plane ICP~\cite{Chen1992} attempts to address this for common cases, but thereby limits the set of objectives which can be minimized, losing the freedom introduced by Neugebauer and Fitzgibbon.
Recently Rusinkiewicz~\cite{Rusinkiewicz2019} presents a symmetric objective function for ICP which is particularly suited to the original task of point set to point set alignment, but not of point set to parametric surface alignment, which is the focus of this paper.
Note that we are explicitly in a non-symmetric scenario. An extension of~\cite{Rusinkiewicz2019} could be considered by sampling the model, but this would lose the structure that the model provides, and which is available in many real scenarios.

Several authors have attempted to address the shortcomings of ICP by estimating unknown correspondences simultaneously with model parameters.
Sullivan~and Ponce~\cite{Sullivan1998} present one of the first systems to do so in a model-fitting context, and Cashman~and Fitzgibbon~\cite{Cashman2013} elaborate on this idea to also fit smooth surface models to silhouette constraints.
Subsequently, Taylor \etal~\cite{Taylor2014, Taylor2016} and Khamis \etal~\cite{Khamis2015} demonstrate the benefits of lifted optimization in the area of articulated hand tracking.
However, all of these methods required complicated smooth surface constructions, in stark contrast to the freedom available when using ICP.
This paper addresses this disadvantage, by showing that lifted optimization is equally applicable to models with extremely simple geometry.

Another line of work performs model fitting without estimating any data correspondences at all.
Taylor \etal~\cite{Taylor2017} follow the same approach originally proposed by Fitzgibbon~\cite{fitzgibbon2003} by using articulated distance fields to find correspondences; this is conceptually similar to an ICP closest-point search, but can be implemented extremely efficiently using graphics hardware.
Mueller \etal~\cite{Mueller2019} use a discriminative network to perform dense correspondence regression, thus allowing the model-fitting stage to proceed under the assumption that the correspondences are fixed.
Neither of these approaches are currently suitable for deployment on low-power devices.

%% file: Method.tex
\section{Method}


\def\numpoints{D}


We describe model-fitting problems in a common framework with the following notation:
\begin{itemize}
\item A 3D surface model $S(\theta) \subset \mathbb{R}^3$ parameterized by a vector $\theta$, which might for example specify rigid pose, shape variations or joint angles,
\item A list of sampled data points $\{ {\mathbf x}_i \}_{i=1}^{\numpoints}$ with estimated data normals $\{ {\mathbf x}^\perp_i \}_{i=1}^{\numpoints}$, 
\item An objective function $E(\theta)$ that penalizes differences between the parameterized model and the observed data,
\item An optimizer that iteratively updates the current hypothesis for $\theta$ to locally minimize the objective function.
\end{itemize}

We follow the model fitting work of Taylor \etal \cite{Taylor2016} which optimizes a differentiable lifted objective function with a Levenberg optimizer.
The smoothness of the subdivision surface model used in \cite{Taylor2016} encourages good convergence properties, but at the cost of expensive function evaluations for the surface positions, normals and their derivatives.
We focus on efficiency and investigate the requirements for surface and normal smoothness in a lifted optimizer.



\subsection{Phong surface model}
\label{subsection:phong}

The key idea is to generate surface normal vectors by linearly interpolating vertex normals over each planar triangle. 
This is the same approximation of a smooth surface used in the Phong shading method for computer graphics rendering~\cite{Phong1975}, motivating our use of the `Phong surface' moniker. 

The surface model is defined by a triangle control mesh containing $N$ control vertices each with a position and normal vector, $V(\theta) \in \mathbb{R}^{6 \times N}$.  
The model also has a fixed triangulation of the vertices, where each triangle in the mesh corresponds to a parameterized triangular patch of the surface.

As illustrated in Fig.~\ref{fig:triangleDomain}, let $u = \{p, v, w\}$ be a surface coordinate where $p \in \mathbb{N}$ is the index of a triangular patch, and $v \in [0,1], w \in [0, 1-v]$ parameterize the unit triangle.
$S({ u}, \theta) \in \mathbb{R}^3$ denotes the surface position and $S^\perp({ u}, \theta) \in \mathbb{R}^3$ the unit-length normal to the surface, both evaluated at the given coordinate ${u}$. 
Let ${\boldsymbol v}_1(\theta), {\boldsymbol v}_2(\theta), {\boldsymbol v}_3(\theta)$ be the control vertex positions and ${\boldsymbol v}^{\perp}_1(\theta),{\boldsymbol v}^{\perp}_2(\theta), {\boldsymbol v}^{\perp}_3(\theta)$ the control vertex normals of the $p$th triangular patch as specified by $u$, where
$\boldsymbol{v}_i(\theta)$ and $\boldsymbol{v}^\perp_i(\theta)$ are determined by the pose and/or shape parameter vector $\theta$.
Then the Phong surface evaluation is defined simply as a linear interpolation of the control vertices:
\begin{align}
S({ u}, \theta) & = (1-v-w){\boldsymbol v}_1(\theta) + v{\boldsymbol v}_2(\theta) + w{\boldsymbol v}_3(\theta), \\
{\boldsymbol c}({ u}, \theta) & = (1-v-w){\boldsymbol v}^{\perp}_1(\theta) + v{\boldsymbol v}^{\perp}_2(\theta) + w{\boldsymbol v}^{\perp}_3(\theta), \\
S^{\perp}({ u}, \theta) & = \frac{{\boldsymbol c}({ u}, \theta) }{\| {\boldsymbol c}({ u}, \theta)  \|},
\end{align}
where ${\boldsymbol c}({ u}, \theta)$  is the interpolated normal direction vector.

We give the partial derivatives with respect to $v, w$ and $\theta$ compactly in terms of the total differentials:
\begin{align*}
\text{d}S(u, \theta) & = \text{d}v({\boldsymbol v}_2 - {\boldsymbol v}_1) + \text{d}w({\boldsymbol v}_3 - {\boldsymbol v}_1) + (1-v-w) \text{d}{\boldsymbol v}_1 + v\text{d}{\boldsymbol v}_2 + w\text{d}{\boldsymbol v}_3, \\
\text{d}{\boldsymbol c}({ u}, \theta) & = \text{d}v({\boldsymbol v}^\perp_2 - {\boldsymbol v}^\perp_1) + \text{d}w({\boldsymbol v}^\perp_3 - {\boldsymbol v}^\perp_1) + (1-v-w) \text{d}{\boldsymbol v}^\perp_1 + v\text{d}{\boldsymbol v}^\perp_2 + w\text{d}{\boldsymbol v}^\perp_3, \\
\text{d}S^\perp(u, \theta) & = \frac{1}{\| {\boldsymbol c} \|}\left(I_3-\frac{{\boldsymbol c} {\boldsymbol c}^T}{\| {\boldsymbol c}\|^2}\right) \text{d}{\boldsymbol c}.
\end{align*}
The partials can be read off from this notation as the coefficient of the relevant differential, e.g. $\tfrac{\partial S}{\partial v} = {\boldsymbol v}_2 - {\boldsymbol v}_1$ by setting $\text{d}v=1, \text{d}w=0, \text{d}{\boldsymbol v}_i = {\bf 0}, \text{d}{\boldsymbol v}^\perp_i = {\bf 0}$ in $\text{d}S$.


\begin{figure}[t]
\centering
 \includegraphics[width=0.6\linewidth]{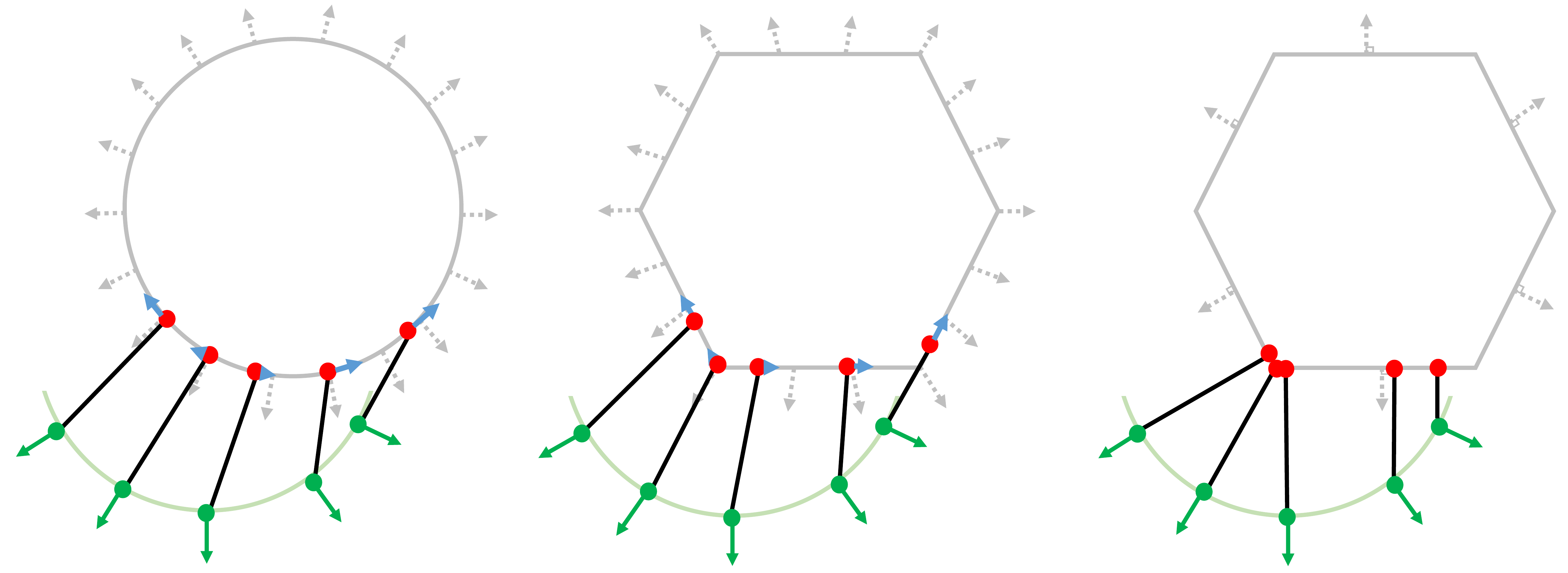} \\
 \begin{subfigure}{0.2\linewidth}
  \caption{Smooth surface}
 \end{subfigure}
 \begin{subfigure}{0.2\linewidth}
  \caption{Phong surface}
 \end{subfigure}
 \begin{subfigure}{0.2\linewidth}
  \caption{Triangular mesh}
 \end{subfigure}
 \caption{Illustration of the cross section of a surface model (gray) with normals (dashed gray) close to some target data (green). With a continuous normal field, either from a smooth model (a) or from a Phong surface (b), the normal term forces shown in blue drive the red correspondences towards their correct location, improving convergence. These forces do not exist for a triangular mesh (c) where the normals are constant on the facet and so the derivatives are all zero.}
 \label{fig:AdvantageOfSmoothNormals}
\end{figure}

\subsection{Lifted optimization with the Phong surface}
\label{subsection:lifted}


The objective function of the optimization in model fitting typically includes data terms and problem-dependent prior terms or regularization terms.
Here we briefly describe the data terms.

We penalize the distance from each data point ${\mathbf x}_i$ to its corresponding surface point defined by the surface coordinate $u_i$, and the difference in surface orientation to the associated data normal~${\mathbf x}^{\perp}_i$, using
\begin{equation}
E(\theta, U) = \frac{1}{\numpoints}\sum_{i=1}^\numpoints \Bigl( {\|S({u_i},\theta) - {\mathbf x}_i\|^2} + \lambda_n {\| S^\perp({ u_i},\theta) - {\mathbf x}^{\perp}_i\|^2} \Bigr)
\label{eq:data}
\end{equation}
where $\numpoints$ is the number of data points, $\lambda_n$ is the contribution weight for normals, and $U = \{  u_i \}_{i=1}^\numpoints$ are the surface coordinates corresponding to each data point.

Note that the surface coordinates $U$ are optimized jointly with $\theta$ by the lifted optimizer, whereas the ICP optimizer alternates between updating $\theta$ and $U$.
This means that the lifted optimizer can choose to slide these coordinates on the surface to better match the data points and data normals, while simultaneously updating the shape or pose hypothesis.
On the other hand, the ICP optimizer operates on only one set of variables each iteration, without access to gradient information in the other variables.
$U$ can be viewed as latent variables as we eventually retain only the final $\theta$ estimate.
As in~\cite{Taylor2016}, the lifted optimizer uses Levenberg steps with damping.

Figure~\ref{fig:AdvantageOfSmoothNormals} illustrates how the continuous surface normals for the subdivision surface and Phong surface causes the data normal term to `pull' the coordinates in the directions of the blue arrows, leading to faster convergence than for the triangular mesh.


Note that \emph{lifted optimization} has a close relation to \emph{point-to-plane ICP} but is mathematically richer while being more efficient.
Both optimizers allow a model to slide against the data in each iteration, improving convergence. 
However, \emph{point-to-plane ICP} addresses this for only one energy formulation (point-to-model distances) whereas a lifted optimizer generalizes to arbitrary differentiable objectives.
For example, we have a normal disparity term in our energy (Eq.|\ref{eq:data}), which is critical for faster convergence as shown in Fig.~\ref{fig:toyExConvergeNormalTerm} and Table~\ref{tab:toyExAccuracyNormalTermLifted};
point-to-plane ICP cannot minimize this objective.
See our supplementary material for more illustrations.


\subsection{Correspondence update on triangles}
\label{sec:corresp}

The surface types \emph{Subdiv.}, \emph{Phong} and \emph{Tri.~mesh} are all defined by a triangular control mesh.
As mentioned in Sec~\ref{subsection:phong}, we write correspondences as surface coordinates $u = \{p, v, w\}$.
The entire surface can be viewed as a collection of triangular patches indexed by $p$, with each patch parameterized by the unit triangle.

\begin{figure}[t]
	\centering
	\begin{subfigure}[b]{0.29\linewidth}
		\centering
		\includegraphics[width=0.9\linewidth]{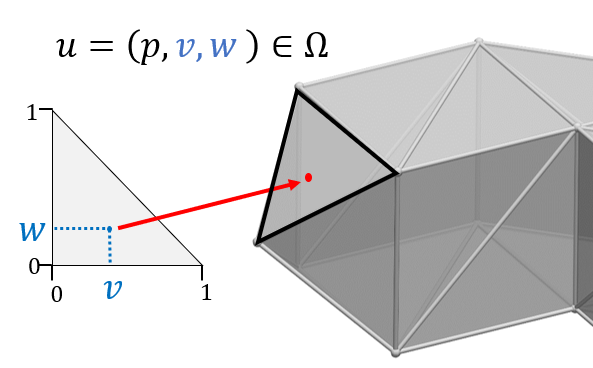}
				\caption{Surface parameterization}
		\label{fig:triangleDomain}
	\end{subfigure}
	\begin{subfigure}[b]{0.5\linewidth}
		\centering
		\includegraphics[width=0.5\linewidth]{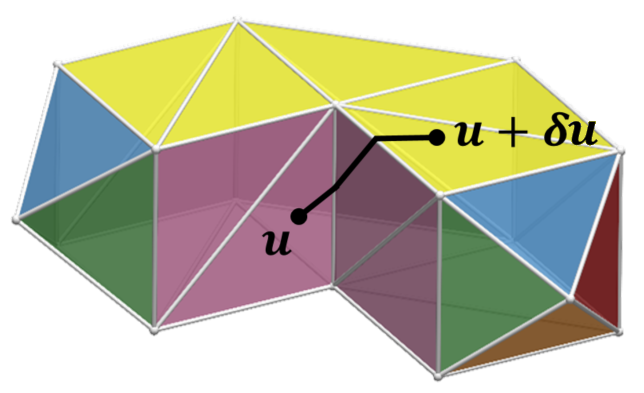}
		\includegraphics[width=0.3\linewidth]{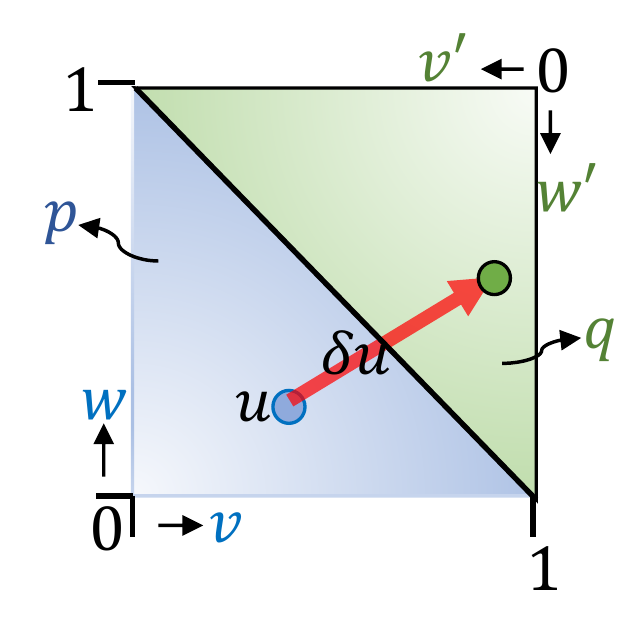}
				\caption{Correspondence update across triangles.}
		\label{fig:triangleWalking}
	\end{subfigure}
	\caption{
		(a) A surface correspondence $u$ that lies on the $p$-th triangle patch, and its coordinate $(v, w)$ in the unit-triangle domain of this patch. (b) Walking on a triangle mesh to apply update $\delta u$ (figure from~\cite{Taylor2014}) and walking in the domain space.}
	\label{fig:triangles}
\end{figure}

After a Levenberg step in lifted optimization, we apply an update $u \vcentcolon = u + \delta u$, which may involve walking across adjacent triangles.
We follow the same triangle-walking scheme as in~\cite{Cashman2013,Taylor2014}, as illustrated in Fig.~\ref{fig:triangleWalking}. 

Fig.~\ref{fig:triangleWalking} (right) shows a single transition of a correspondence from one triangle to its neighbour;
$u$ and $\delta u$ are denoted as a 2D point and a 2D vector respectively, in the domain space of the current patch $p$ (colored in light blue).
When $u + \delta u$ leaves the domain of triangle $p$, we calculate the partial update such that $u + r \delta u$ lies on the boundary of that triangle, and then map the remaining update $(1 - r) \delta u$ to the adjacent patch $q$ (colored in light green).
For \emph{Subdiv.}, this results in a path that is tangent-continuous on the surface because this surface has $C^1$ continuity across patches.
Note that this $C^1$ assumption does not hold for \emph{Phong} and \emph{Tri.\ mesh}; however, we implement the walking in exactly the same way for all surface types and find that it works well in practice.

%% file: Experiments.tex
\section{Experiments}

We compare the three surface types mentioned above: Loop subdivision surface (\emph{Subdiv}), Phong surface (\emph{Phong}) and triangular mesh surface (\emph{Tri. mesh}), in both \emph{Lifted} and \emph{ICP} optimization frameworks.
First we analyze their efficiency and convergence properties on rigid pose estimation of an ellipsoid.
We further extend them to a more challenging scenario: fully articulated hand tracking.
Specifically, we implement and evaluate these methods in the hand tracker by Taylor \etal~\cite{Taylor2016} and in a hand tracker for HoloLens 2~\cite{HoloLens2019}.

In each experiment, we define the control parameters of each surface as follows: for each control vertex of \emph{Subdiv}, we compute its
position and normal on the limit Loop subdivision surface; we use these limit positions as the vertices of \emph{Tri. mesh},
and use both limit positions and normals to define control vertices and normals of \emph{Phong}.
Note that this definition is purely to give surface models that are comparable for \emph{Phong}, \emph{Subdiv} and \emph{Tri. mesh}, and we are free to define the Phong Surface model (i.e.\ control vertex positions and normals) in the way that best represents the target geometry.

We run most experiments on a desktop machine equipped with an Intel\textsuperscript{\textregistered} Xeon\textsuperscript{\textregistered} W-2155 CPU and 32GB RAM, except for Fig.~\ref{subfig:hololens-perf-accuracy}, which is on a Microsoft HoloLens 2.

%
%
%

\subsection{Rigid pose alignment of an ellipsoid}
~\label{sec:toyEx}

\noindent {\bf Problem.} 
We parametrize the ellipsoid pose as a 6D vector $\theta$, storing translation and (axis-angle) rotation $[t_x, t_y, t_z, r_x, r_y, r_z]$.
It defines a translation vector $\mathbf{t}(\theta) \in \mathbb{R}^3$ and a rotation matrix $R(\theta) \in \mathbb{R}^{3 \times 3}$.
Given a template mesh, each control vertex position ${\boldsymbol v}_i$ is posed according to $\theta$:
\begin{align}
{\boldsymbol v}_i(\theta) = R(\theta){\boldsymbol v}_i + {\boldsymbol t}(\theta).
\end{align}
For Phong surfaces, we express also control vertex normals as a function of $\theta$ (this is more efficient than re-computing the normals from the posed vertices):
\begin{align}
{\boldsymbol v}^{\perp}_i(\theta) =  R(\theta) {\boldsymbol v}^{\perp}_i.
\end{align}
This defines the ${\boldsymbol v}_i(\theta)$, ${\boldsymbol v}^{\perp}_i(\theta)$ introduced in Sec.~\ref{subsection:phong}.
Derivatives $\frac{\partial {\boldsymbol v}_i(\theta)}{\partial \theta}, \frac{\partial {\boldsymbol v}^{\perp}_i(\theta)}{\partial \theta}$ can be trivially computed.

The objective function includes only the data term stated in Eq.~\ref{eq:data}. 
The surface evaluations are computed as in Sec.~\ref{subsection:phong}.
In our experiments, we set $\numpoints = 200$. 
For fair comparisons among various surfaces and especially \emph{Tri. mesh}, we did parameter sweeping on $\lambda_n$ in the range $[0.0, 1.0]$ for fitting 400 random rotations. 
Based on the average fitting error, we find optimal $\lambda_n = 1.0$ for \emph{Subdiv} and \emph{Phong}, and $\lambda_n = 0.05$ for \emph{Tri. mesh}.
For the following experiments, for each surface type, we use its optimal $\lambda_n$.
Note that we have included $0.0$ in our $\lambda_n$ sweeping.

\begin{figure}[t]
	\begin{center}
		\includegraphics[width=0.15\linewidth]{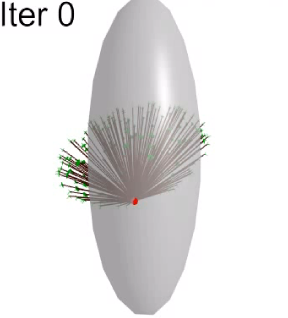}
		\includegraphics[width=0.15\linewidth]{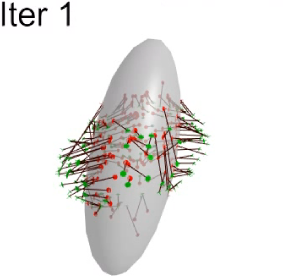}
		\includegraphics[width=0.15\linewidth]{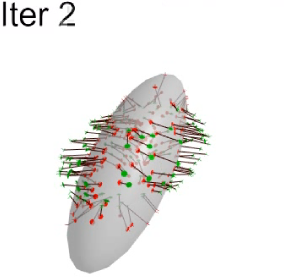}
		\includegraphics[width=0.15\linewidth]{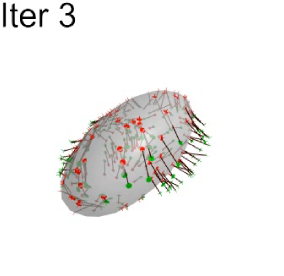}
		\includegraphics[width=0.15\linewidth]{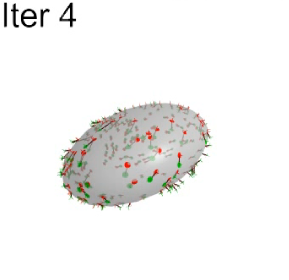}
		\includegraphics[width=0.15\linewidth]{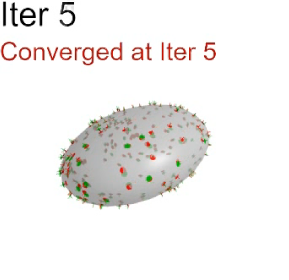}
	\end{center}
	\caption{Lifted optimization using a Phong surface ellipsoid model. Green points are the target data, with black lines joining them to their corresponding surface points in red.}
	\label{fig:ellipsoidConvergence}
\end{figure}

Fig.~\ref{fig:ellipsoidConvergence} shows an example fitting result. See supplementary material for more qualitative comparisons.

\vspace{5pt}
\noindent {\bf Input data.} 
Starting with the axis-aligned ellipsoid (with radii 1, 2, 3) centered at the origin (referred to as \emph{neutral pose}), we apply a target rigid transform to obtain a ground-truth (target) pose.
We then sample randomly 200 data points and normals on the subdivision surface defined by this mesh. 
For quantitative convergence analysis, we only sample from the triangle patches facing the positive z-axis: this gives us an incomplete set of data points. 
We add random noise to the data points and normals by sampling uniform random distributions with range [0.0, 0.1] in each dimension.
The initial starting pose for model fitting is always the neutral pose.

\vspace{5pt}
\noindent {\bf Metrics.} 
For the quantitative analysis, we focus on rotations. 
We compute a pose estimation error by applying the fitted and ground-truth rigid transformations to a fixed vector (e.g., $[1,0,0]$), and measuring the angle (in degrees) between the two transformed vectors.
To allow for the $180^{\circ}$ symmetry of an ellipsoid, we define the fitting error as the minimum of two angles, one computed using $[1,0,0]$ as the fixed vector and the other computed using $[-1,0,0]$ as the fixed vector.

 \begin{figure*}[t]
\input{figs/ToyEx/ellipsoidAccuracyBestF320.tex}
\caption{Accuracy results for the rigid pose alignment of an ellipsoid with \emph{320} facets. Optimizer run for max. 50 iterations.}
\label{fig:toyExAccuracyF320}
\end{figure*}
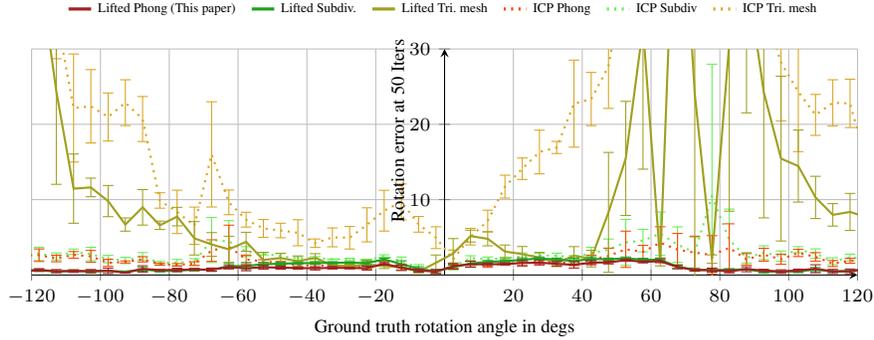

 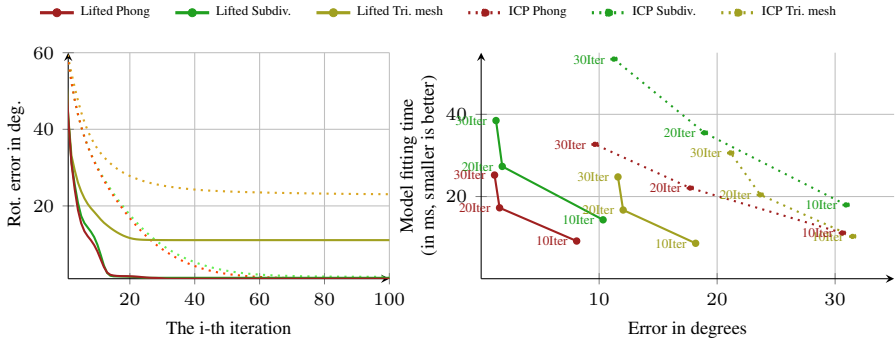
\begin{figure}[t]
\input{figs/ToyEx/ellipsoidConverge.tex}
\caption{Left: Convergence results for the rigid pose alignment of an ellipsoid with 320 facets.
Right: Speed (model-fitting time in miliseconds) vs. accuracy (avg. error) comparisons on lifted and ICP optimizations, for max.10, 20, and 30 iters. Closer to the origin is better.}
\label{fig:toyExConverge}
\end{figure}

\vspace{5pt}
\noindent {\bf Quantitative analysis.} 
We run $400$ trials targeting ground-truth poses $[0, 0, 0, y, y, y]$, where $y$ is uniformly sampled from $(-\pi, \pi)$.
Fig.~\ref{fig:toyExAccuracyF320} and~\ref{fig:toyExConverge} show the performance obtained for the rigid alignment of an ellipsoid model with 320 facets.
Fig.~\ref{fig:toyExAccuracyF320} shows that \emph{Phong} performs as well as \emph{Subdiv.} in accuracy, and much better than \emph{Tri. mesh}, with either \emph{ICP} or \emph{lifted} optimization.
In Fig.~\ref{fig:toyExConverge} (left), we see that \emph{lifted} optimization converges much faster than \emph{ICP} for both \emph{Phong} and \emph{Subdiv}.

Fig.~\ref{fig:toyExConverge} (right) further plots accuracy (on the x-axis) against speed (on the y-axis, measured in milliseconds).
It shows that \emph{Phong} achieves the same level of accuracy as \emph{Subdiv.} and runs as fast as \emph{Tri. mesh} for both \emph{lifted} (solid lines) and \emph{ICP} (dashed lines).
It also shows that lifted optimization converges much faster than ICP, e.g.\ \emph{Lifted Phong} can achieve average rotation error $<10^{\circ}$ within 8 iters, while \emph{ICP Phong} needs 30 iterations.

Note that our runtime for \emph{ICP} is slightly slower than \emph{lifted} here. While a faster ICP might be achievable by further code optimization, we emphasize that the per-iteration cost of lifted is actually comparable to ICP when counting the theoretical FLOP computations. See our supplementary material (part 3) for details.


\vspace{5pt}
\noindent {\bf Data normal term.}
\begin{table}[t]
  \centering
  \begin{tabular}{ l| c| c} 
\hline
Surface type
 &  Avg.~rot.~err. at 10 iters
 &  Avg.~rot.~err. at 50 iters \\
\hline
Subdiv. & $9.89^\circ$ & $1.21^\circ$  \\
\hline
Subdiv.~w/o normal&  $14.81^\circ$ & $2.57^\circ$  \\
\hline
Phong  & ${\bf 8.13^\circ}$   &  ${\bf 0.99^\circ}$ \\
\hline
Phong w/o normal&   $23.24^\circ$  &  $3.54^\circ$ \\
\hline
Tri.~mesh  &  $17.47^\circ$ &  ${ 11.07^\circ}$ \\
\hline
Tri.~mesh w/o normal & $23.24^\circ$ &  $3.54^\circ$ \\
\hline
 \end{tabular}
   \caption{Average rotation error after 10 and 50 {lifted optimization} iterations, with and without data normal term.}
\label{tab:toyExAccuracyNormalTermLifted}
\end{table}
Here we assess the importance of the data normal term in Eq.~\ref{eq:data}.
In particular, we demonstrate that this term is critical for fast convergence in both lifted and ICP optimizations, and that a continuous normal field improves both the basin of convergence and the accuracy of pose estimation.
%

 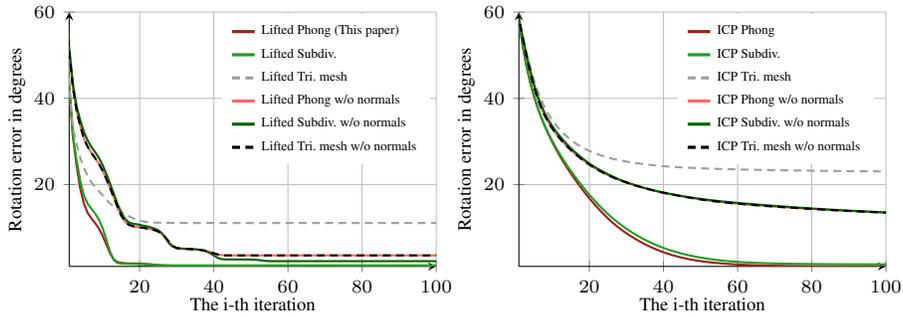
\begin{figure}[t]
\input{figs/ToyEx/ellipsoidConvergeNormalTerm.tex}
\caption{Impact of the data normal term on optimization convergence, for the rigid pose alignment of an ellipsoid (320 facets). 
Left:  Lifted optimization. Right: ICP optimization.
}
\label{fig:toyExConvergeNormalTerm}
\end{figure}

As shown in Fig.~\ref{fig:toyExConvergeNormalTerm}, \emph{Phong} and \emph{Subdiv.}, which have a continuous normal field, converge much faster when the normal term is included, and achieve better accuracy after convergence (see also Table~\ref{tab:toyExAccuracyNormalTermLifted}).
Note that {Phong w/o normal} (solid bright red)  coincides with {Tri.~mesh w/o normal} (dashed black) for both lifted and ICP optimization.

For \emph{Tri.~mesh} (dashed lines in Fig.~\ref{fig:toyExConvergeNormalTerm} (left), we observe faster convergence with the data normal term within 10 iterations, but the accuracy reached after convergence is worse.
We believe this is because 
the piecewise-constant surface normals introduce local minima for the correspondences, where reassignment to a different triangle causes discrete jumps in the energy, and it is difficult for the optimizer to find a global minimum for these correspondences as the partial derivatives $\frac{\partial \tilde{S}^\perp}{\partial u}$ are zero.
Note that the accuracy of \emph{Tri.~mesh w/o normal} is still 3 times worse than the \emph{Phong} and \emph{Subdiv.} models when including the normal term (see Table~\ref{tab:toyExAccuracyNormalTermLifted}, third column).


\subsection{Performance on hand tracking}

\begin{figure}[t]
\begin{center}
\includegraphics[width=0.18\linewidth]{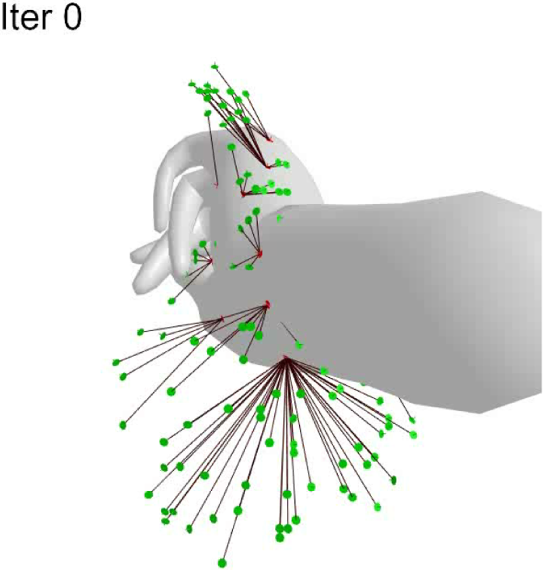}
\includegraphics[width=0.18\linewidth]{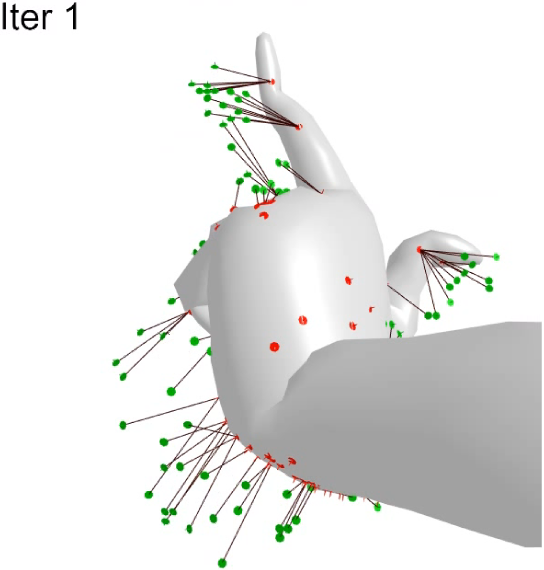}
\includegraphics[width=0.18\linewidth]{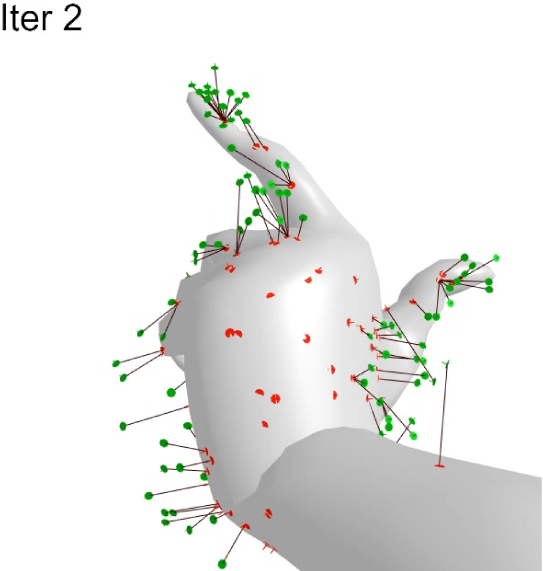}
\includegraphics[width=0.18\linewidth]{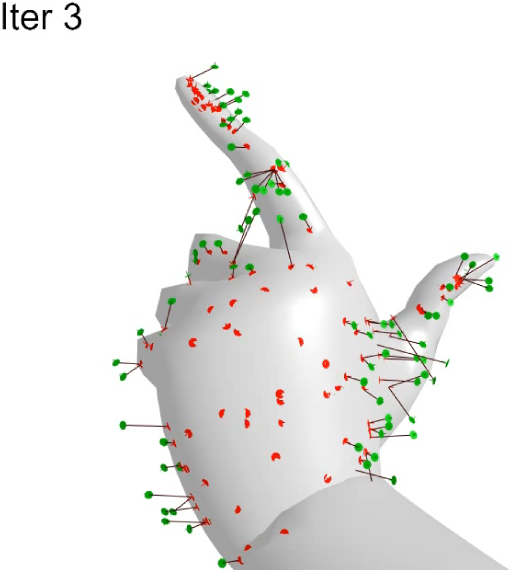}
\includegraphics[width=0.18\linewidth]{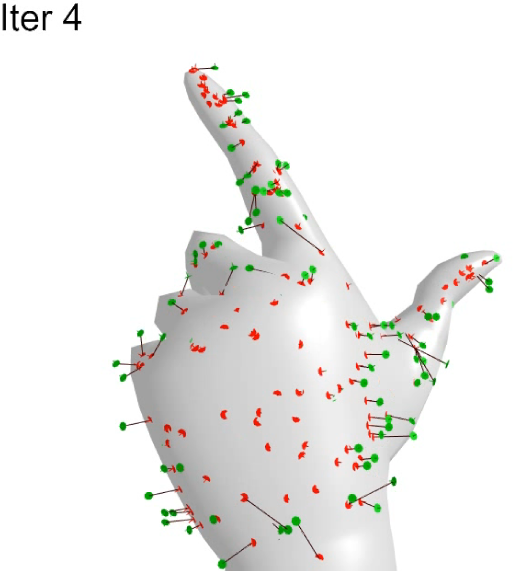}
\end{center}
   \caption{A starting point and four iterations of lifted optimization using a Phong surface hand model. Green points are the target data, with black lines joining them to their corresponding surface points in red.}
\label{fig:handConvergence}
\end{figure}

The \emph{lifted} optimizer used in the hand tracker by Taylor \etal~\cite{Taylor2016} uses Loop subdivision surfaces.
In this experiment, we simply replace their subdivision surface with our Phong surface. We leave everything else unchanged.
Fig.~\ref{fig:handConvergence} shows an example result obtained fitting our Phong surface (in Fig.~\ref{subfig:surfaces-phong}) with lifted optimization.

\begin{figure}[t]
	\begin{center}
		\includegraphics[width=0.48\linewidth]{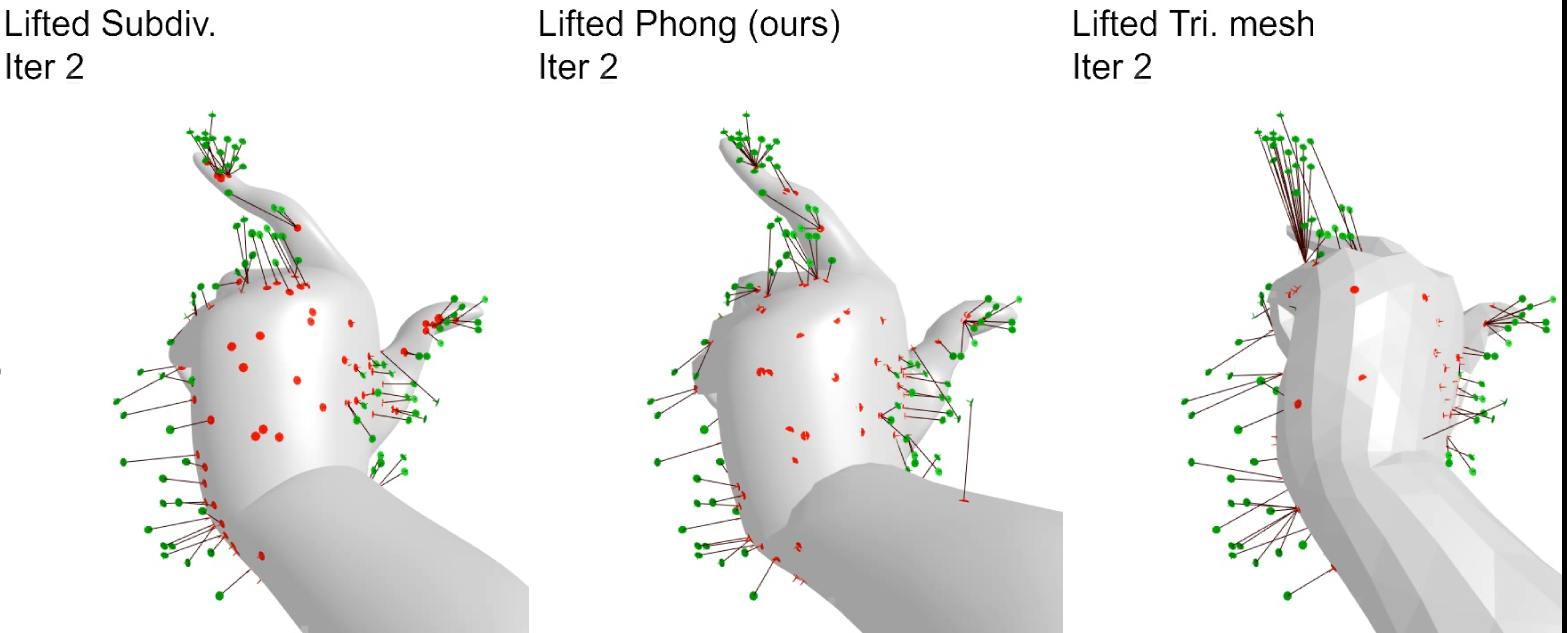}
		\includegraphics[width=0.48\linewidth]{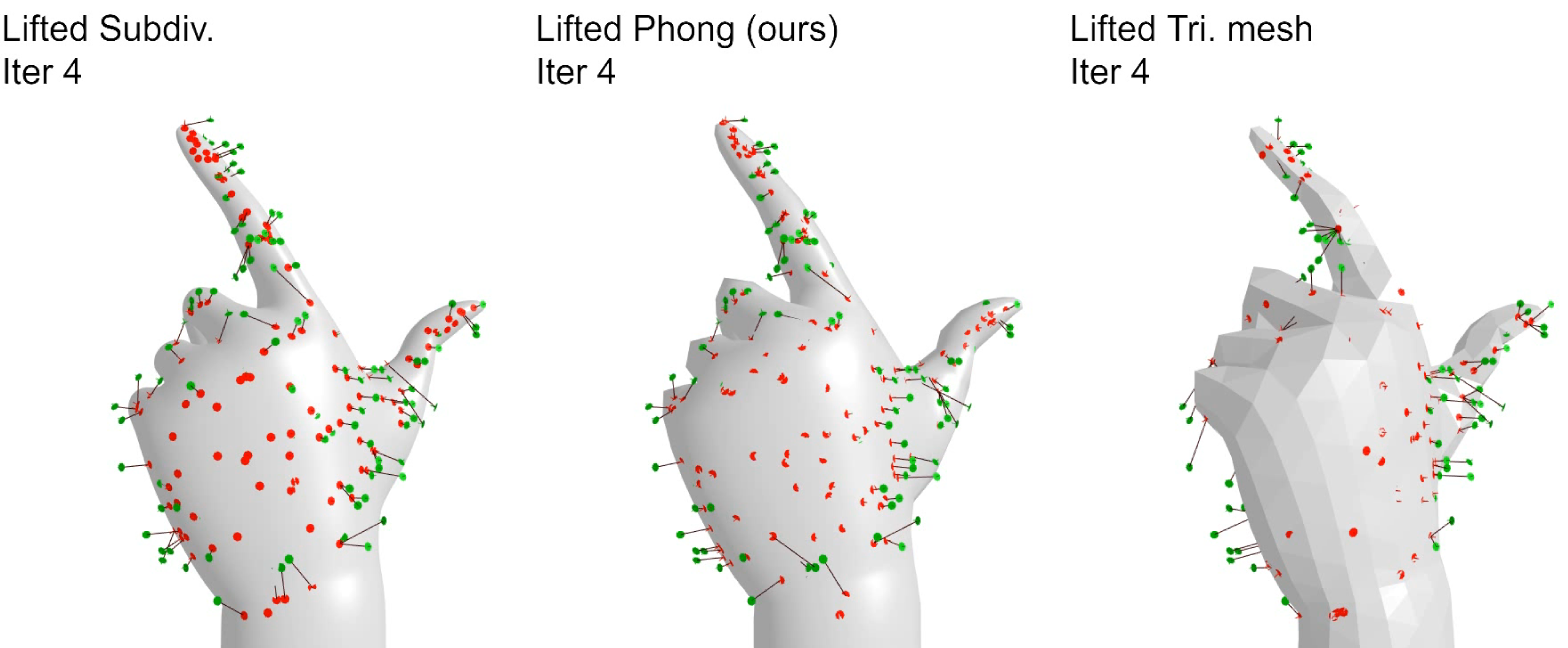}
	\end{center}
	\caption{Fitted hand model result at the 2nd and 4th iterations of lifted and ICP optimization using various surface types.}
	\label{fig:handConvergenceCmp}
\end{figure}
Fig.~\ref{fig:handConvergenceCmp} compares qualitatively the hand model fitting results at some iterations of lifted and ICP optimization using various surface types, given the same starting point. See the supplementary video for all iterations.

\noindent {\bf Problem.} 
We optimize a set of hand pose parameters $\theta \in \mathbb{R}^{28}$; $\theta$ stores hand orientation and translation, plus the 22 joint angles of the hand skeleton.
From $\theta$, the hand surface vertices are computed by Linear Blend Skinning~\cite{Khamis2015}.

For Phong surface, the 3D mesh vertex normals are deformed in the same way as vertex positions according to their LBS weights.
This computation for the control normals gives a close approximation of the normals that we would obtain by rederiving normals from the posed positions of local vertices, with far greater efficiency.

We refer the reader to~\cite{Taylor2016} for details on the objective function, data and experimental setup and evaluation metrics.

 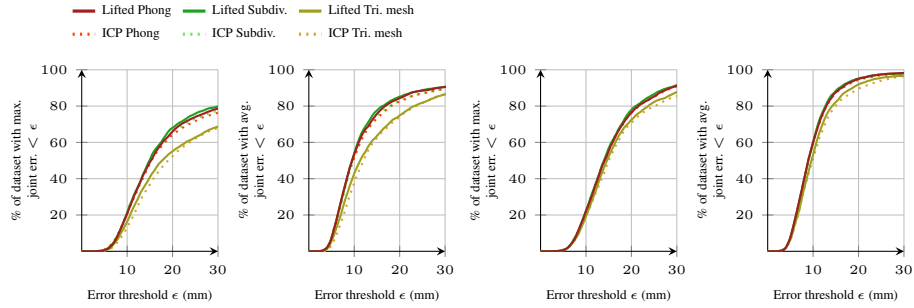
\begin{figure}[t]
	\input{figs/ModelOpt/DexterAccuracyMethods.tex}
	\caption{Accuracy comparison of surface types with lifted optimizer and ICP on Dexter.}\label{fig:dexter-accuracy}
	\label{fig:dexter-accuracy}
\end{figure}
 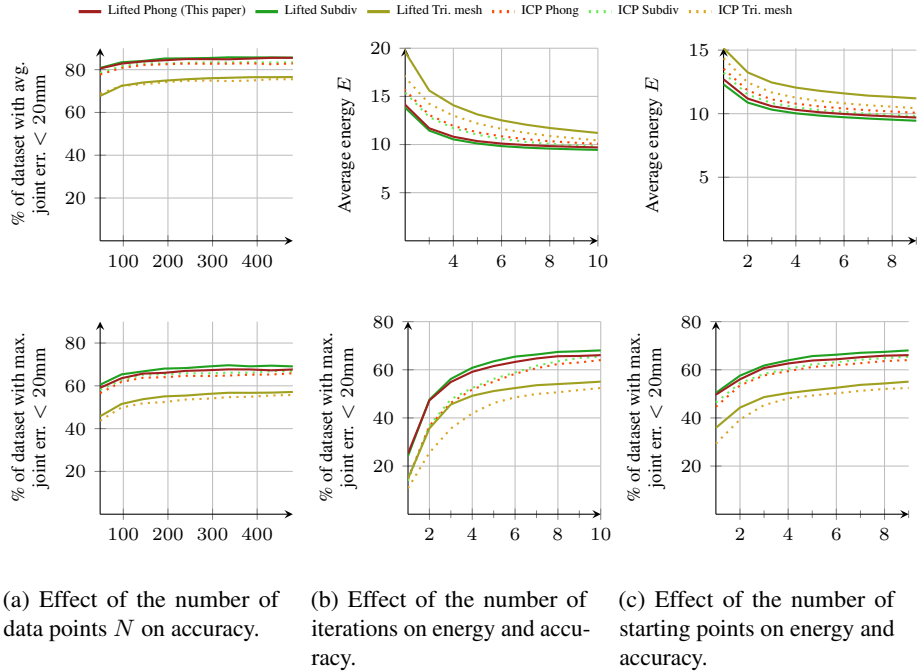
\begin{figure}[t]
	\input{figs/ModelOpt/DexterEfficiency.tex}
	\caption{Effect of alternative optimization configurations when fitting to {Dexter}. The results showed are for per-frame model fitting.}
	\label{fig:dexter-efficiency}
\end{figure}

\noindent {\bf Performance on Dexter.} In Fig.~\ref{fig:dexter-accuracy}, \ref{fig:dexter-efficiency} and~\ref{subfig:dexter-perf-accuracy}, we compare the accuracy and the computational efficiency of the tracker with different surface types on the Dexter dataset~\cite{Sridhar2013}. We adopt the same experimental setup as in~\cite{Taylor2016}.


The tracking settings for Fig.~\ref{fig:dexter-accuracy} are 192 data points, 10 starting points and 10 iterations. 
Fig.~\ref{subfig:dexter-accuracy-perframe} shows the max and average error of per-frame fitting, i.e.\ fitting each frame independently with 10 new starting points.
Fig.~\ref{subfig:dexter-accuracy-tracking} shows the max and average error of tracking, i.e.\ using the tracked pose in previous frame (if available) as one of the 10 starting points for current frame.
Both show that on this dataset, \emph{Phong} performs as well as \emph{Subdiv} and achieves higher accuracy than \emph{Tri. mesh}.
\emph{Lifted} optimization is slightly better than \emph{ICP}.


\noindent {\bf Robustness to initialization.} 
Model-fitting methods are often sensitive to initialization, due to the non-convex objectives. 
This is why the tracking accuracy (Fig.~\ref{subfig:dexter-accuracy-tracking}) is better than the per-frame fitting (Fig.~\ref{subfig:dexter-accuracy-perframe}), where the optimization starts from scratch each time.
The gap between \emph{Phong} and \emph{Tri. mesh} is larger in the per-frame 
fitting (Fig.~\ref{subfig:dexter-accuracy-perframe}), which means that the smooth normal field is the key for faster convergence when the starting point is poorer.
We emphasize the importance of fast convergence in live experience, as tracking failures often occur when hands out of the field of view, or in the presence of self- and object-occlusions.

The computational cost of model fitting is dominated by 3 variables: (i) the number of data points in the data term; (ii) the number of starting points used to initialize the optimizer; and (iii) the number of iterations for each starting point. 
Fig.~\ref{fig:dexter-efficiency} shows the impact of these variables on accuracy and convergence. 
Again, \emph{Phong} behaves similarly to \emph{Subdiv.}, and much better than \emph{Tri. mesh}.
Fig.~\ref{subfig:dexter-efficiency-iterations} shows how \emph{Lifted} exhibits better convergence than \emph{ICP}, confirming the conclusion in~\cite{Taylor2016}.

These tests show that the smooth normal field of the model is important for faster convergence in lifted optimization; it is not actually relevant whether the mesh geometry is smooth or not.

Fig.~\ref{subfig:dexter-perf-accuracy} shows the speed vs. accuracy plot in the per-frame fitting case (192 data points, 10 starting points).
For each surface type, the number of iterations varies from 2 to 10.
The x-axis reports the accuracy, measured as the percentage of dataset frames that have average joint error $<20mm$. 
The y-axis reports the speed in FPS (per starting point) of the model-fitting stage, i.e.\ not including the preprocessing time. 
For example, in the \emph{lifted} case (solid lines), if we require to run the fitting at 50fps, we can perform 6 iterations for \emph{Phong} and \emph{Tri. mesh}, but only 4 iterations for \emph{Subdiv.}, and \emph{Phong} provides the highest accuracy at this speed.
Alternatively, if we require the fitter to reach near $80\%$ accuracy, we can run 4 iterations with \emph{Phong} and \emph{Subdiv.}, but \emph{Phong} is $20\%$ faster.
So \emph{Phong} achieves almost the same level of accuracy as \emph{Subdiv}, while being as cheap as \emph{Tri. mesh} in terms of efficiency.

Similar conclusions can be drawn for ICP optimizations (dashed lines in Fig.~\ref{subfig:dexter-perf-accuracy}).
As pointed out earlier at the end of Sec.~\ref{sec:toyEx}, our runtime for \emph{ICP} is slightly slower than \emph{lifted}, but we show that per-iteration cost of lifted is actually comparable with ICP in our supplementary material, part 3.

Note that the accuracy achieved by \emph{Lifted Subdiv.} after 8 iterations coincides with that achieved by \emph{Lifted Phong} after 10 iterations.
This is because \emph{Lifted Phong} already converged after 8 iterations, and further iterations do not improve accuracy further (see also Fig.~\ref{subfig:dexter-efficiency-iterations}).

 \begin{figure}
	\input{figs/ModelOpt/DexterPerfAccuracy.tex}
	\caption{(a) The speed (model-fitting speed in fps for one starting point) vs. accuracy (percentage of frames with avg. joint error less than 20 mm) for per-frame fitting using the hand tracker by Taylor \etal~\cite{Taylor2016} on Dexter. Upwards and to the right is better. The dots from top to bottom on each line denote the number of iterations being 2, 4, 6, 8, 10.
(b) The speed vs. accuracy for our hand tracker on HoloLens 2.   }
	\label{fig:hand-perf-accuracy}
\end{figure}
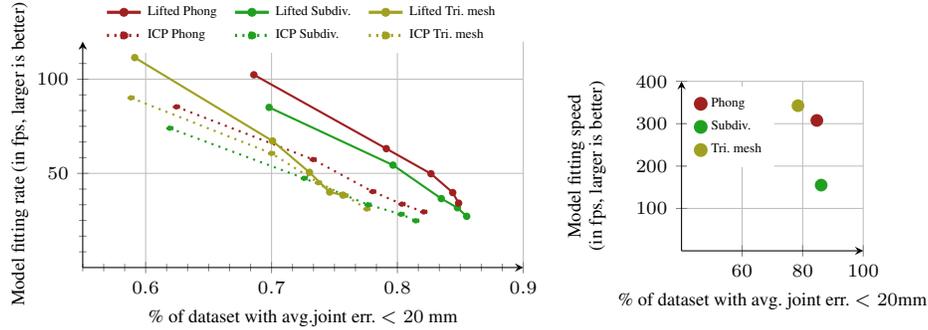

\noindent {\bf Performance on HoloLens 2.} 
Starting from the work of~\cite{Taylor2016}, we made many improvements to enable us to run a hand tracker in real time on the HoloLens 2~\cite{HoloLens2019}, a mobile device with very limited computational and power resources. The Phong surface model presented here was one of the key efficiency improvements that was required. 
Fig.~\ref{subfig:hololens-perf-accuracy} shows the speed vs. accuracy of various surface types with lifted optimization in this hand tracker on HoloLens 2, evaluated on a captured depth dataset. 
The \emph{Phong} surface (red dot) can be evaluated twice as fast as the subdivision surface (green dot), and gives the same level of accuracy.


%% file: figs/ToyEx/ellipsoidAccuracyBestF320.tex
%

\let \yaxiswidth \undefined
\newlength{\yaxiswidth}
\setlength{\yaxiswidth}{0.35\columnwidth}

\begin{minipage}[b]{\textwidth}
\begin{tikzpicture}
	\begin{axis}[
		xlabel={Ground truth rotation angle in degs}, 
		ylabel={Rotation error at 50 Iters},
		ylabel style={align=flush right,text width=\yaxiswidth,yshift=-0.15cm,xshift=-0.5cm},
		xticklabel={\pgfmathparse{1*\tick}$\pgfmathprintnumber{\pgfmathresult}$},
		yticklabel={\pgfmathparse{1*\tick}$\pgfmathprintnumber{\pgfmathresult}$},
		width=0.9\columnwidth,
		height=0.8\plotheight,
		ymin=0,
		ymax=30,
		xmin =-120,
		xmax=120,
		legend style={anchor=south,at={(0.5, 1.1)},/tikz/every even column/.append style={column sep=0.01\columnwidth}},
		legend columns=6,
		reverse legend
		]
\addplot+[chira3b, dotted, error bars/.cd, y fixed, y dir=both, y explicit, error bar style={solid}] table[x=x, y=y,y error=error] {figs/ToyEx/Data/ellipsoid_F320/bestLambdas/ICP_PieceWiseLinear_err_2pi_iter50.data}; \addlegendentry{ICP Tri. mesh}
\addplot[chira2b, dotted, error bars/.cd, y fixed, y dir=both, y explicit, error bar style={solid}] table[x=x, y=y,y error=error] {figs/ToyEx/Data/ellipsoid_F320/bestLambdas/ICP_ApproxLoopTriangleMonomialDeg4_err_2pi_iter50.data}; \addlegendentry{ICP Subdiv}
\addplot[chira1b, dotted, error bars/.cd, y fixed, y dir=both, y explicit, error bar style={solid}] table[x=x, y=y,y error=error] {figs/ToyEx/Data/ellipsoid_F320/bestLambdas/ICP_PhongSurface_err_2pi_iter50.data}; \addlegendentry{ICP Phong}

\addplot[chira3, error bars/.cd, y fixed, y dir=both, y explicit, error bar style={solid}] table[x=x, y=y,y error=error] {figs/ToyEx/Data/ellipsoid_F320/bestLambdas/Joint_PieceWiseLinear_err_2pi_iter50.data}; \addlegendentry{Lifted Tri. mesh}
\addplot[chira2, error bars/.cd, y fixed, y dir=both, y explicit, error bar style={solid}] table[x=x, y=y,y error=error] {figs/ToyEx/Data/ellipsoid_F320/bestLambdas/Joint_ApproxLoopTriangleMonomialDeg4_err_2pi_iter50.data}; \addlegendentry{Lifted Subdiv.}
\addplot[chira1, error bars/.cd, y fixed, y dir=both, y explicit, error bar style={solid}] table[x=x, y=y,y error=error] {figs/ToyEx/Data/ellipsoid_F320/bestLambdas/Joint_PhongSurface_err_2pi_iter50.data}; \addlegendentry{Lifted Phong (This paper)}
	
	\end{axis}
\end{tikzpicture}
\end{minipage}
\vskip-3pt

%% file: figs/ToyEx/ellipsoidConverge.tex
%


\let \yaxiswidth \undefined
\newlength{\yaxiswidth}
\setlength{\yaxiswidth}{0.35\columnwidth}

\begin{minipage}[b]{\columnwidth}
	\begin{tikzpicture}
	\begin{axis}[
	xlabel={The i-th iteration},
	ylabel={Rot. error in deg.},
	ylabel style={align=flush center,text width=\yaxiswidth,yshift=-0.1cm},
	xticklabel={\pgfmathparse{1*\tick}$\pgfmathprintnumber{\pgfmathresult}$},
	yticklabel={\pgfmathparse{1*\tick}$\pgfmathprintnumber{\pgfmathresult}$},
	width=0.35\columnwidth,
	height=0.8\plotheight,
	ymax=60,
	xmax=100,
	]	
	\addplot[chira3b, dotted] table[header=false] {figs/ToyEx/Data/ellipsoid_F320/lambdaBest_Iter100/ICP_PieceWiseLinear_converge_per_iter_angle_err.data}; 
	\addplot[chira2b, dotted] table[header=false] {figs/ToyEx/Data/ellipsoid_F320/lambdaBest_Iter100/ICP_ApproxLoopTriangleMonomialDeg4_converge_per_iter_angle_err.data};
	\addplot[chira1b, dotted] table[header=false] {figs/ToyEx/Data/ellipsoid_F320/lambdaBest_Iter100/ICP_PhongSurface_converge_per_iter_angle_err.data};
	\addplot[chira3] table[header=false] {figs/ToyEx/Data/ellipsoid_F320/lambdaBest_Iter100/Joint_PieceWiseLinear_converge_per_iter_angle_err.data}; 
	\addplot[chira2] table[header=false] {figs/ToyEx/Data/ellipsoid_F320/lambdaBest_Iter100/Joint_ApproxLoopTriangleMonomialDeg4_converge_per_iter_angle_err.data}; 
	\addplot[chira1] table[header=false] {figs/ToyEx/Data/ellipsoid_F320/lambdaBest_Iter100/Joint_PhongSurface_converge_per_iter_angle_err.data}; 
	\end{axis}
	\hfil
	\hfil
	\begin{axis}[
	nodes near coords,
	point meta=explicit symbolic,
xlabel={Error in degrees},
ylabel={Model fitting time \\ (in ms, smaller is better)},	
ylabel style={align=flush center,text width=\yaxiswidth,yshift=-0.1cm},
width=0.45\columnwidth,
height=0.8\plotheight,
xshift=0.45\columnwidth,
ymin=0,
ymax=55,
xmin=0,
xmax =35,
legend image post style={scale=1.5,mark options={mark size=1pt, line width=0.2pt}},
legend style={anchor=south,at={(-0.1, 1.1)},/tikz/every even column/.append style={column sep=0.02\columnwidth}},
legend columns=6,
reverse legend	
]
\addplot[every node near coord/.append style={ anchor=east,font=\tiny},chira3, dotted, mark=otimes*, mark options={scale=0.5,draw=chira3,fill=chira3}]  table[header=true, meta=config] {figs/ToyEx/err_time/ToyAvgVsTime-ICP-PiecewiseLinear-Perframe.data}; \addlegendentry{ICP Tri. mesh}
\addplot+[every node near coord/.append style={anchor=east,font=\tiny},chira2, dotted,  mark=otimes*, mark options={scale=0.5,draw=chira2,fill=chira2}] table[header=true, meta=config] {figs/ToyEx/err_time/ToyAvgVsTime-ICP-ApproxLoop-Perframe.data}; \addlegendentry{ICP Subdiv.}	
\addplot+[every node near coord/.append style={ anchor=east,font=\tiny}, chira1, dotted,  mark=otimes*,  mark options={scale=0.5,draw=chira1,fill=chira1}]  table[header=true, meta=config] {figs/ToyEx/err_time/ToyAvgVsTime-ICP-Phong-Perframe.data};	\addlegendentry{ICP Phong}		
\addplot+[every node near coord/.append style={anchor=east,font=\tiny},chira3, mark=otimes*, mark options={scale=0.5,draw=chira3,fill=chira3}]  table[header=true, meta=config] {figs/ToyEx/err_time/ToyAvgVsTime-Joint-PiecewiseLinear-Perframe.data}; \addlegendentry{Lifted Tri. mesh}
\addplot+[every node near coord/.append style={ anchor=east,font=\tiny},chira2,  mark=otimes*, mark options={scale=0.5,draw=chira2,fill=chira2}] table[header=true, meta=config] {figs/ToyEx/err_time/ToyAvgVsTime-Joint-ApproxLoop-Perframe.data}; \addlegendentry{Lifted Subdiv.}	
\addplot+[every node near coord/.append style={ anchor=east,font=\tiny}, chira1, mark=otimes*,  mark options={scale=0.5,draw=chira1,fill=chira1}]  table[header=true, meta=config] {figs/ToyEx/err_time/ToyAvgVsTime-Joint-Phong-Perframe.data};	\addlegendentry{Lifted Phong}
\end{axis}	
	\end{tikzpicture}
\end{minipage}
\vskip-6pt

%% file: figs/ToyEx/ellipsoidConvergeNormalTerm.tex
%

\let \yaxiswidth \undefined
\newlength{\yaxiswidth}
\setlength{\yaxiswidth}{0.35\columnwidth}

\begin{minipage}[b]{\columnwidth}
	\begin{tikzpicture}
	\begin{axis}[
	xlabel={The i-th iteration},
	ylabel={Rotation error in degrees},
	xlabel style={align=flush center, yshift=0.15cm},
	ylabel style={align=flush center,text width=\yaxiswidth,yshift=-0.1cm},
	xticklabel={\pgfmathparse{1*\tick}$\pgfmathprintnumber{\pgfmathresult}$},
	yticklabel={\pgfmathparse{1*\tick}$\pgfmathprintnumber{\pgfmathresult}$},
	width=0.4\columnwidth,
	height=0.9\plotheight,
	ymax=60,
	xmax=100,
	legend style={anchor=south,at={(0.7, 0.4)},/tikz/every even column/.append style={column sep=0.01\columnwidth}},
	legend columns=1
	]
	\addplot[chira1] table[header=false] {figs/ToyEx/Data/ellipsoid_F320/lambdaBest_Iter100/Joint_PhongSurface_converge_per_iter_angle_err.data}; \addlegendentry{Lifted Phong (This paper)}
	\addplot[chira2] table[header=false] {figs/ToyEx/Data/ellipsoid_F320/lambdaBest_Iter100/Joint_ApproxLoopTriangleMonomialDeg4_converge_per_iter_angle_err.data};\addlegendentry{Lifted Subdiv.}
	\addplot[chira3c, ldotted] table[header=false] {figs/ToyEx/Data/ellipsoid_F320/lambdaBest_Iter100/Joint_PieceWiseLinear_converge_per_iter_angle_err.data}; \addlegendentry{Lifted Tri. mesh}
	\addplot[chira1c] table[header=false] {figs/ToyEx/Data/ellipsoid_F320/lambda0_Iter100/Joint_PhongSurface_converge_per_iter_angle_err.data}; \addlegendentry{Lifted Phong w/o normals}
	\addplot[chira2c] table[header=false] {figs/ToyEx/Data/ellipsoid_F320/lambda0_Iter100/Joint_ApproxLoopTriangleMonomialDeg4_converge_per_iter_angle_err.data}; \addlegendentry{Lifted Subdiv. w/o normals}
	\addplot[chira3d, ldotted] table[header=false] {figs/ToyEx/Data/ellipsoid_F320/lambda0_Iter100/Joint_PieceWiseLinear_converge_per_iter_angle_err.data}; \addlegendentry{Lifted Tri. mesh w/o normals}	
	\end{axis}
	\begin{axis}[
	xlabel={The i-th iteration},
	ylabel={Rotation error in degrees},
	xlabel style={align=flush center, yshift=0.15cm},
	ylabel style={align=flush center,text width=\yaxiswidth,yshift=-0.1cm},
	xticklabel={\pgfmathparse{1*\tick}$\pgfmathprintnumber{\pgfmathresult}$},
	yticklabel={\pgfmathparse{1*\tick}$\pgfmathprintnumber{\pgfmathresult}$},
	xshift=0.49\columnwidth,
	width=0.4\columnwidth,
	height=0.9\plotheight,
	ymax=60,
	xmax=100,
	legend style={anchor=south,at={(0.7, 0.4)},/tikz/every even column/.append style={column sep=0.01\columnwidth}},
	legend columns=1
	]
	
	\addplot[chira1] table[header=false] {figs/ToyEx/Data/ellipsoid_F320/lambdaBest_Iter100/ICP_PhongSurface_converge_per_iter_angle_err.data}; \addlegendentry{ICP Phong}
	\addplot[chira2] table[header=false] {figs/ToyEx/Data/ellipsoid_F320/lambdaBest_Iter100/ICP_ApproxLoopTriangleMonomialDeg4_converge_per_iter_angle_err.data}; \addlegendentry{ICP Subdiv.}
	\addplot[chira3c, ldotted] table[header=false] {figs/ToyEx/Data/ellipsoid_F320/lambdaBest_Iter100/ICP_PieceWiseLinear_converge_per_iter_angle_err.data}; \addlegendentry{ICP Tri. mesh}
	\addplot[chira1c] table[header=false] {figs/ToyEx/Data/ellipsoid_F320/lambda0_Iter100/ICP_PhongSurface_converge_per_iter_angle_err.data}; \addlegendentry{ICP Phong w/o normals}
	\addplot[chira2c] table[header=false] {figs/ToyEx/Data/ellipsoid_F320/lambda0_Iter100/ICP_ApproxLoopTriangleMonomialDeg4_converge_per_iter_angle_err.data}; \addlegendentry{ICP Subdiv. w/o normals}
	\addplot[chira3d, ldotted] table[header=false] {figs/ToyEx/Data/ellipsoid_F320/lambda0_Iter100/ICP_PieceWiseLinear_converge_per_iter_angle_err.data}; \addlegendentry{ICP Tri. mesh w/o normals}
	\end{axis}
	\end{tikzpicture}
\end{minipage}

%% file: figs/ModelOpt/DexterAccuracyMethods.tex

\let \yaxiswidth \undefined
\newlength{\yaxiswidth}
\setlength{\yaxiswidth}{0.35\columnwidth}

\begin{minipage}[t]{0.495\columnwidth}
	\begin{tikzpicture}
		\begin{axis}[
		xlabel={Error threshold $\epsilon$ (mm)},
		ylabel={\maxmetriclabel{\epsilon}},		
		xticklabel={\pgfmathparse{100*\tick}$\pgfmathprintnumber{\pgfmathresult}0$},
		yticklabel={\pgfmathparse{100*\tick}$\pgfmathprintnumber{\pgfmathresult}$},
		xlabel style={font=\tiny},		
		ylabel style={align=flush center,text width=\yaxiswidth,yshift=-0.2cm,font=\tiny},		
		yticklabel style={font=\tiny},
		xticklabel style={font=\tiny},
		width=0.3\columnwidth,
		height=0.4\columnwidth,
		ymax=1,
		xmax=0.03,
		legend style={anchor=south,at={(1.2, 1.1)},/tikz/every even column/.append style={column sep=0.01\columnwidth}},
		legend columns=3,
		reverse legend
		]
		\addplot[chira3b,dotted] table[header=false] {figs/Data/DexterMaxResults-PiecewiseIcp-f98e07-PerFrame.data};\addlegendentry{ICP Tri. mesh}
		\addplot[chira2b,dotted] table[header=false] {figs/Data/DexterMaxResults-LoopIcp-f98e07-PerFrame.data}; \addlegendentry{ICP Subdiv.}
		\addplot[chira1b,dotted] table[header=false] {figs/Data/DexterMaxResults-PhongIcp-f98e07-PerFrame.data};\addlegendentry{ICP Phong}	
		\addplot[chira3] table[header=false] {figs/Data/DexterMaxResults-PiecewiseJoint-f98e07-PerFrame.data};\addlegendentry{Lifted Tri. mesh}
		\addplot[chira2] table[header=false] {figs/Data/DexterMaxResults-LoopJoint-f98e07-PerFrame.data}; \addlegendentry{Lifted Subdiv.}	
		\addplot[chira1] table[header=false] {figs/Data/DexterMaxResults-PhongJoint-f98e07-PerFrame.data}; 	\addlegendentry{Lifted Phong}		
		\end{axis}	
		\begin{axis}[
		xlabel={Error threshold $\epsilon$ (mm)},
		ylabel={\averagemetriclabel{\epsilon}},
		xticklabel={\pgfmathparse{100*\tick}$\pgfmathprintnumber{\pgfmathresult}0$},
		yticklabel={\pgfmathparse{100*\tick}$\pgfmathprintnumber{\pgfmathresult}$},
		xlabel style={font=\tiny},		
ylabel style={align=flush center,text width=\yaxiswidth,yshift=-0.2cm,font=\tiny},		
yticklabel style={font=\tiny},
xticklabel style={font=\tiny},
width=0.3\columnwidth,
height=0.4\columnwidth,
		xshift=0.5\columnwidth,		
		ymax=1,
		xmax=0.03,
		legend style={anchor=south,at={(1.2, 1.1)},/tikz/every even column/.append style={column sep=0.01\columnwidth}},
		legend columns=3,
		reverse legend
		]
		\addplot[chira3b,dotted] table[header=false] {figs/Data/DexterAvgResults-PiecewiseIcp-f98e07-PerFrame.data};
		\addplot[chira2b,dotted] table[header=false] {figs/Data/DexterAvgResults-LoopIcp-f98e07-PerFrame.data}; 
		\addplot[chira1b,dotted] table[header=false] {figs/Data/DexterAvgResults-PhongIcp-f98e07-PerFrame.data}; 
		\addplot[chira3] table[header=false] {figs/Data/DexterAvgResults-PiecewiseJoint-f98e07-PerFrame.data};
		\addplot[chira2] table[header=false] {figs/Data/DexterAvgResults-LoopJoint-f98e07-PerFrame.data}; 
		\addplot[chira1] table[header=false] {figs/Data/DexterAvgResults-PhongJoint-f98e07-PerFrame.data}; 		
		\end{axis}
	\end{tikzpicture}\vskip-6pt
           \captionsetup{font=small,labelfont=small}
	\subcaption{ Max (left) and Avg. (right) joint error on \emph{per-frame fitting}. }\label{subfig:dexter-accuracy-perframe}
\end{minipage}
\hfill
\begin{minipage}[t]{0.495\columnwidth}	
	\begin{tikzpicture}
	\begin{axis}[
	xlabel={Error threshold $\epsilon$ (mm)},
	ylabel={\maxmetriclabel{\epsilon}},
	xticklabel={\pgfmathparse{100*\tick}$\pgfmathprintnumber{\pgfmathresult}0$},
	yticklabel={\pgfmathparse{100*\tick}$\pgfmathprintnumber{\pgfmathresult}$},
		xlabel style={font=\tiny},		
ylabel style={align=flush center,text width=\yaxiswidth,yshift=-0.2cm,font=\tiny},		
yticklabel style={font=\tiny},
xticklabel style={font=\tiny},
width=0.3\columnwidth,
height=0.4\columnwidth,
	ymax=1,
	xmax=0.03
	]
	\addplot[chira3b,dotted] table[header=false] {figs/Data/DexterMaxResults-PiecewiseIcp-f98e07-tracking.data}; 
	\addplot[chira2b,dotted] table[header=false] {figs/Data/DexterMaxResults-LoopIcp-f98e07-tracking.data};
	\addplot[chira1b,dotted] table[header=false] {figs/Data/DexterMaxResults-PhongIcp-f98e07-tracking.data}; 
	\addplot[chira3] table[header=false] {figs/Data/DexterMaxResults-PiecewiseJoint-f98e07-tracking.data};%
	\addplot[chira2] table[header=false] {figs/Data/DexterMaxResults-LoopJoint-f98e07-tracking.data}; 
	\addplot[chira1] table[header=false] {figs/Data/DexterMaxResults-PhongJoint-f98e07-tracking.data};%
	\end{axis}		
		\begin{axis}[
		xlabel={Error threshold $\epsilon$ (mm)},
		ylabel={\averagemetriclabel{\epsilon}},
		xticklabel={\pgfmathparse{100*\tick}$\pgfmathprintnumber{\pgfmathresult}0$},
		yticklabel={\pgfmathparse{100*\tick}$\pgfmathprintnumber{\pgfmathresult}$},
		xlabel style={font=\tiny},		
ylabel style={align=flush center,text width=\yaxiswidth,yshift=-0.2cm,font=\tiny},		
yticklabel style={font=\tiny},
xticklabel style={font=\tiny},
width=0.3\columnwidth,
height=0.4\columnwidth,
		xshift=0.5\columnwidth,
		ymax=1,
		xmax=0.03
		]
		\addplot[chira3b,dotted] table[header=false] {figs/Data/DexterAvgResults-PiecewiseIcp-f98e07-tracking.data}; 
		\addplot[chira2b,dotted] table[header=false] {figs/Data/DexterAvgResults-LoopIcp-f98e07-tracking.data};
		\addplot[chira1b,dotted] table[header=false] {figs/Data/DexterAvgResults-PhongIcp-f98e07-tracking.data}; 
		\addplot[chira3] table[header=false] {figs/Data/DexterAvgResults-PiecewiseJoint-f98e07-tracking.data};%
		\addplot[chira2] table[header=false] {figs/Data/DexterAvgResults-LoopJoint-f98e07-tracking.data}; 
		\addplot[chira1] table[header=false] {figs/Data/DexterAvgResults-PhongJoint-f98e07-tracking.data};%
		\end{axis}
\end{tikzpicture}\vskip-6pt
\captionsetup{font=small,labelfont=small}
\subcaption{Max (left) and Avg. (right) joint error on \emph{tracking}.}\label{subfig:dexter-accuracy-tracking}
\end{minipage}

%% file: figs/ModelOpt/DexterEfficiency.tex
%

\let \yaxiswidth \undefined
\newlength{\yaxiswidth}
\setlength{\yaxiswidth}{0.35\columnwidth}

\begin{minipage}[t]{0.3\columnwidth}
	\begin{tikzpicture}	
	\begin{axis}[
	ylabel={\averagemetriclabel{20\text{mm}}},
	ylabel style={align=flush center,text width=\yaxiswidth,yshift=-0.15cm},
	yticklabel={\pgfmathparse{100*\tick}$\hphantom{1}\pgfmathprintnumber{\pgfmathresult}$},
	width=0.7\columnwidth,
	height=0.7\columnwidth,
	ymin=0,
	ymax=0.9,
	legend style={anchor=south, at={(1.75, 1.1)},/tikz/every even column/.append style={column sep=0.01\columnwidth}},
	legend columns=6,
	reverse legend	
	]
	\addplot[chira3b,dotted] table[header=false] {figs/ModelOpt/Dexter-PerFrame-Sweeps/dexter_avg_acc_piece_icp_pixels_to_sample_at_20mm.dat};\addlegendentry{ICP Tri. mesh}
	\addplot[chira2b,dotted] table[header=false] {figs/ModelOpt/Dexter-PerFrame-Sweeps/dexter_avg_acc_loop_icp_pixels_to_sample_at_20mm.dat};\addlegendentry{ICP Subdiv}
	\addplot[chira1b,dotted] table[header=false] {figs/ModelOpt/Dexter-PerFrame-Sweeps/dexter_avg_acc_phong_icp_pixels_to_sample_at_20mm.dat};\addlegendentry{ICP Phong}
	\addplot[chira3] table[header=false] {figs/ModelOpt/Dexter-PerFrame-Sweeps/dexter_avg_acc_piece_joint_pixels_to_sample_at_20mm.dat};\addlegendentry{Lifted Tri. mesh}
	\addplot[chira2] table[header=false] {figs/ModelOpt/Dexter-PerFrame-Sweeps/dexter_avg_acc_loop_joint_pixels_to_sample_at_20mm.dat};\addlegendentry{Lifted Subdiv}
	\addplot[chira1] table[header=false] {figs/ModelOpt/Dexter-PerFrame-Sweeps/dexter_avg_acc_phong_joint_pixels_to_sample_at_20mm.dat};\addlegendentry{Lifted Phong (This paper)}
	\end{axis}
	\end{tikzpicture}
	\vskip-25pt
\begin{tikzpicture}
	\begin{axis}[
	ylabel={\maxmetriclabel{20\text{mm}}},
	ylabel style={align=flush center,text width=\yaxiswidth,yshift=-0.15cm},
	yticklabel={\pgfmathparse{100*\tick}$\hphantom{1}\pgfmathprintnumber{\pgfmathresult}$},
	width=0.7\columnwidth,
height=0.7\columnwidth,
	ymin=0,
	ymax=0.9
	]
	\addplot[chira3b,dotted] table[header=false] {figs/ModelOpt/Dexter-PerFrame-Sweeps/dexter_max_acc_piece_icp_pixels_to_sample_at_20mm.dat};
\addplot[chira2b,dotted] table[header=false] {figs/ModelOpt/Dexter-PerFrame-Sweeps/dexter_max_acc_loop_icp_pixels_to_sample_at_20mm.dat};
\addplot[chira1b,dotted] table[header=false]
{figs/ModelOpt/Dexter-PerFrame-Sweeps/dexter_max_acc_phong_icp_pixels_to_sample_at_20mm.dat};
	\addplot[chira3] table[header=false] {figs/ModelOpt/Dexter-PerFrame-Sweeps/dexter_max_acc_piece_joint_pixels_to_sample_at_20mm.dat};
	\addplot[chira2] table[header=false] {figs/ModelOpt/Dexter-PerFrame-Sweeps/dexter_max_acc_loop_joint_pixels_to_sample_at_20mm.dat};
	\addplot[chira1] table[header=false]
	{figs/ModelOpt/Dexter-PerFrame-Sweeps/dexter_max_acc_phong_joint_pixels_to_sample_at_20mm.dat};
	\end{axis}
	\end{tikzpicture} \vskip-6pt
 	\captionsetup{font=small,labelfont=small}
	\subcaption{Effect of the number of data points $N$ on accuracy.\newline }\label{subfig:dexter-efficiency-numdata}
\end{minipage}
\hfil
\begin{minipage}[t]{0.3\columnwidth}
	\begin{tikzpicture}
	\begin{axis}[
	minor xtick={3,5,7,9},
	ylabel style={align=flush center,text width=\yaxiswidth,yshift=-0.15cm},
	yticklabel={$\hphantom{1}\pgfmathprintnumber{\tick}$},
	ylabel={\\[3.7pt]Average energy $E$},
	legend pos=north east,
	width=0.7\columnwidth,
height=0.7\columnwidth,
	ymin=0,
	ymax=20,
	]
	\addplot[chira3b,dotted] table[header=false,y error index=2] {figs/ModelOpt/Dexter-PerFrame-Sweeps/dexter_energy_piece_icp_iterations.dat}; 
\addplot[chira2b,dotted] table[header=false,y error index=2] {figs/ModelOpt/Dexter-PerFrame-Sweeps/dexter_energy_loop_icp_iterations.dat}; 
\addplot[chira1b,dotted] table[header=false,y error index=2] {figs/ModelOpt/Dexter-PerFrame-Sweeps/dexter_energy_phong_icp_iterations.dat}; 
	\addplot[chira3] table[header=false,y error index=2] {figs/ModelOpt/Dexter-PerFrame-Sweeps/dexter_energy_piece_joint_iterations.dat}; 
	\addplot[chira2] table[header=false,y error index=2] {figs/ModelOpt/Dexter-PerFrame-Sweeps/dexter_energy_loop_joint_iterations.dat}; 
	\addplot[chira1] table[header=false,y error index=2] {figs/ModelOpt/Dexter-PerFrame-Sweeps/dexter_energy_phong_joint_iterations.dat}; 
	\end{axis}
	\end{tikzpicture}
	\vskip-25pt	
	\begin{tikzpicture}
	\begin{axis}[
	minor xtick={3,5,7,9},
	ylabel={\maxmetriclabel{20\text{mm}}},
	ylabel style={align=flush center,text width=\yaxiswidth,yshift=-0.15cm},
	yticklabel={\pgfmathparse{100*\tick}$\hphantom{1}\pgfmathprintnumber{\pgfmathresult}$},
	width=0.7\columnwidth,
height=0.7\columnwidth,
	ymin=0.0,
	ymax=0.8
	]
	\addplot[chira3b,dotted] table[header=false] {figs/ModelOpt/Dexter-PerFrame-Sweeps/dexter_max_acc_piece_icp_iterations_at_20mm.dat};%
\addplot[chira2b,dotted] table[header=false] {figs/ModelOpt/Dexter-PerFrame-Sweeps/dexter_max_acc_loop_icp_iterations_at_20mm.dat};%
\addplot[chira1b,dotted] table[header=false] {figs/ModelOpt/Dexter-PerFrame-Sweeps/dexter_max_acc_phong_icp_iterations_at_20mm.dat};%
	\addplot[chira3] table[header=false] {figs/ModelOpt/Dexter-PerFrame-Sweeps/dexter_max_acc_piece_joint_iterations_at_20mm.dat};%
	\addplot[chira2] table[header=false] {figs/ModelOpt/Dexter-PerFrame-Sweeps/dexter_max_acc_loop_joint_iterations_at_20mm.dat};%
	\addplot[chira1] table[header=false] {figs/ModelOpt/Dexter-PerFrame-Sweeps/dexter_max_acc_phong_joint_iterations_at_20mm.dat};
	\end{axis}
	\end{tikzpicture}\vskip-6pt
	 \captionsetup{font=small,labelfont=small}
	\subcaption{Effect of the number of iterations on energy and accuracy.}\label{subfig:dexter-efficiency-iterations}
\end{minipage}
\hfil
\begin{minipage}[t]{0.3\columnwidth}
	\begin{tikzpicture}
	\begin{axis}[
	minor xtick={3,5,7,9},
	ylabel style={align=flush center,text width=\yaxiswidth,yshift=-0.15cm},
	yticklabel={$\hphantom{10}\pgfmathprintnumber{\tick}$},
	ylabel={\\[3.7pt]Average energy $E$},
	legend pos=north east,
	width=0.7\columnwidth,
height=0.7\columnwidth,
	ymin=0,
	]
	\addplot[chira3b,dotted] table[header=false,y error index=2] {figs/ModelOpt/Dexter-PerFrame-Sweeps/dexter_energy_piece_icp_startpoints.dat}; 
\addplot[chira2b,dotted] table[header=false,y error index=2] {figs/ModelOpt/Dexter-PerFrame-Sweeps/dexter_energy_loop_icp_startpoints.dat}; 	
\addplot[chira1b,dotted] table[header=false,y error index=2] {figs/ModelOpt/Dexter-PerFrame-Sweeps/dexter_energy_phong_icp_startpoints.dat};	
	\addplot[chira3] table[header=false,y error index=2] {figs/ModelOpt/Dexter-PerFrame-Sweeps/dexter_energy_piece_joint_startpoints.dat}; 
	\addplot[chira2] table[header=false,y error index=2] {figs/ModelOpt/Dexter-PerFrame-Sweeps/dexter_energy_loop_joint_startpoints.dat}; 	
	\addplot[chira1] table[header=false,y error index=2] {figs/ModelOpt/Dexter-PerFrame-Sweeps/dexter_energy_phong_joint_startpoints.dat}; 
	\end{axis}
	\end{tikzpicture}
		\vskip-25pt
	\begin{tikzpicture}
	\begin{axis}[
	minor xtick={3,5,7,9},
	ylabel={\maxmetriclabel{20\text{mm}}},
	ylabel style={align=flush center,text width=\yaxiswidth,yshift=-0.15cm},
	yticklabel={\pgfmathparse{100*\tick}$\hphantom{1}\pgfmathprintnumber{\pgfmathresult}$},
	width=0.7\columnwidth,
height=0.7\columnwidth,
	ymin=0,
	ymax=0.8,
	set layers=standard,
	every axis plot/.append style={
		on layer={axis foreground}
	}
	]
	\addplot[chira3b,dotted] table[header=false] {figs/ModelOpt/Dexter-PerFrame-Sweeps/dexter_max_acc_piece_icp_startpoints_at_20mm.dat};%
\addplot[chira2b,dotted] table[header=false] {figs/ModelOpt/Dexter-PerFrame-Sweeps/dexter_max_acc_loop_icp_startpoints_at_20mm.dat};%
\addplot[chira1b,dotted] table[header=false] {figs/ModelOpt/Dexter-PerFrame-Sweeps/dexter_max_acc_phong_icp_startpoints_at_20mm.dat};%
	\addplot[chira3] table[header=false] {figs/ModelOpt/Dexter-PerFrame-Sweeps/dexter_max_acc_piece_joint_startpoints_at_20mm.dat};%
	\addplot[chira2] table[header=false] {figs/ModelOpt/Dexter-PerFrame-Sweeps/dexter_max_acc_loop_joint_startpoints_at_20mm.dat};%
	\addplot[chira1] table[header=false] {figs/ModelOpt/Dexter-PerFrame-Sweeps/dexter_max_acc_phong_joint_startpoints_at_20mm.dat};%
	\end{axis}
	\end{tikzpicture}\vskip-6pt
	 \captionsetup{font=small,labelfont=small}
	\subcaption{Effect of the number of starting points on energy and accuracy.}\label{subfig:dexter-efficiency-startpoints}
\end{minipage}

%% file: figs/ModelOpt/DexterPerfAccuracy.tex
%

\let \yaxiswidth \undefined
\newlength{\yaxiswidth}
\setlength{\yaxiswidth}{0.35\columnwidth}

\begin{minipage}[t]{0.6\columnwidth}
	\begin{tikzpicture}
	\begin{axis}[
	minor tick num=5,
	xlabel={\averagemetriclabel{20} mm},
	ylabel={Model fitting rate (in fps, larger is better)},	
	ylabel style={align=flush center,text width=\yaxiswidth,yshift=-0.1cm},
	width=0.8\columnwidth,
	height=0.8\plotheight,	
	ymin=0,
	ymax=120,
	xmin=0.55,
	xmax =0.9,
	legend image post style={scale=1.5,mark options={mark size=1pt, line width=0.2pt}},
	legend style={anchor=south,at={(0.5, 0.95)},/tikz/every even column/.append style={column sep=0.02\columnwidth}},
	legend columns=3,
	reverse legend	
	]
	\addplot[every node near coord/.append style={xshift=-14pt, anchor=west,font=\tiny},chira3, dotted, mark=otimes*, mark options={scale=0.5,draw=chira3,fill=chira3}]  table[header=true] {figs/ModelOpt/Dexter-PerFrame-Sweeps/DexterAvgVsTime-icp-PiecewiseLinear-Perframe.data}; \addlegendentry{ICP Tri. mesh}
	\addplot+[every node near coord/.append style={xshift=-14pt, anchor=west,font=\tiny},chira2, dotted,  mark=otimes*, mark options={scale=0.5,draw=chira2,fill=chira2}] table[header=true] {figs/ModelOpt/Dexter-PerFrame-Sweeps/DexterAvgVsTime-icp-ApproxLoop-Perframe.data}; \addlegendentry{ICP Subdiv.}	
	\addplot+[every node near coord/.append style={xshift=3pt, anchor=west,font=\tiny}, chira1, dotted,  mark=otimes*,  mark options={scale=0.5,draw=chira1,fill=chira1}]  table[header=true] {figs/ModelOpt/Dexter-PerFrame-Sweeps/DexterAvgVsTime-icp-Phong-Perframe.data};	\addlegendentry{ICP Phong}		
	\addplot+[every node near coord/.append style={xshift=-14pt, anchor=west,font=\tiny},chira3, mark=otimes*, mark options={scale=0.5,draw=chira3,fill=chira3}]  table[header=true] {figs/ModelOpt/Dexter-PerFrame-Sweeps/DexterAvgVsTime-PiecewiseLinear-Perframe.data}; \addlegendentry{Lifted Tri. mesh}
	\addplot+[every node near coord/.append style={xshift=-14pt, anchor=west,font=\tiny},chira2,  mark=otimes*, mark options={scale=0.5,draw=chira2,fill=chira2}] table[header=true] {figs/ModelOpt/Dexter-PerFrame-Sweeps/DexterAvgVsTime-ApproxLoop-Perframe.data}; \addlegendentry{Lifted Subdiv.}	
	\addplot+[every node near coord/.append style={xshift=3pt, anchor=west,font=\tiny}, chira1, mark=otimes*,  mark options={scale=0.5,draw=chira1,fill=chira1}]  table[header=true] {figs/ModelOpt/Dexter-PerFrame-Sweeps/DexterAvgVsTime-Phong-Perframe.data};	\addlegendentry{Lifted Phong}		
	\end{axis}
	\end{tikzpicture}\vskip-3pt
           \captionsetup{font=small,labelfont=small}
	\subcaption{Application on the hand tracker by Taylor \etal~\cite{Taylor2016}. }\label{subfig:dexter-perf-accuracy}
\end{minipage}
\begin{minipage}[t]{0.33\columnwidth}
	\begin{tikzpicture}
	\begin{axis}[
	xlabel={\% of dataset with avg. joint err. $< 20$mm},
	ylabel={Model fitting speed \\ (in fps, larger is better)},	
	ylabel style={align=flush center,text width=\yaxiswidth,yshift=-0.1cm},
	yticklabel={$\hphantom{10}\pgfmathprintnumber{\tick}$},
	legend pos=north west,
	width=0.6\columnwidth,
	height=0.6\plotheight,
	ymin=0,
	ymax =400,
	xmin=40,
	xmax = 100,
	]	
	\addplot[only marks, mark options={draw=chira1,fill=chira1},]  table[header=false] {figs/HoloLensDatasetResult/Data/PhongSurfacePerf.dat};  \addlegendentry{Phong}
	\addplot[only marks, mark options={draw=chira2,fill=chira2},] table[header=false] {figs/HoloLensDatasetResult/Data/ApproxLoopPerf.dat};  \addlegendentry{Subdiv.}
	\addplot[only marks, mark options={draw=chira3,fill=chira3},]  table[header=false] {figs/HoloLensDatasetResult/Data/PlanarTrianglesPerf.dat}; \addlegendentry{Tri. mesh }
	\end{axis}
	\end{tikzpicture}\vskip-3pt
	           \captionsetup{font=small,labelfont=small}
	\subcaption{Our hand tracker on HoloLens 2. }
	\label{subfig:hololens-perf-accuracy}
\end{minipage}

%% file: Conclusion.tex

\section{Conclusions}

In this paper we demonstrated that the convergence benefits of lifted optimization are available to a wider range of surface models than was previously thought.
We introduced Phong surfaces, and showed that they provide sufficient information about the local model geometry to allow a model-fitting optimizer to converge fast, whilst requiring a fraction of the compute of expensive smooth surface models.
Beside rigid pose alignment and hand tracking, the proposed method can be applied to various 3D surface model-fitting applications, for example, 3D pose and shape estimation of SMPL body~\cite{Bogo2016}, face~\cite{Li2017} and animals~\cite{ZuffiKJB17,ZuffiKB18}, and is particularly valuable when computational budget is limited.

Given the generalization we show in this paper, a natural question is `what are the set of requirements in order for lifted optimization to work effectively?'
We hypothesize that the only requirement is for a model to provide sufficient approximations to the energy tangent space $\partial E/\partial \theta$ for a gradient-based optimizer to take efficient steps in each iteration, with an implied freedom on global topology and connectivity, as well as the form taken by those approximations.
We intend to explore this hypothesis more fully in future work.


%% file: PhongSurfaceSupplementary.tex
\clearpage
\begin{center}
	\textbf{\large The Phong Surface: Efficient 3D Model Fitting using Lifted Optimization \\ -- Supplemental Materials}
\end{center}
\setcounter{equation}{0}
\setcounter{figure}{0}
\setcounter{table}{0}
\setcounter{section}{0}
\setcounter{page}{1}
\makeatletter
\renewcommand{\theequation}{S\arabic{equation}}
\renewcommand{\thefigure}{S\arabic{figure}}
\renewcommand{\thesection}{S-\Roman{section}}

\section{Lifting vs.\ point-to-plane ICP}
\label{section:point-to-plane-icp}

Here we show that lifted optimization is related to point-to-plane ICP, but is mathematically richer.

\begin{figure}
	\centering
	\begin{subfigure}[b]{0.37\linewidth}
		\centering
		\includegraphics[width=\linewidth]{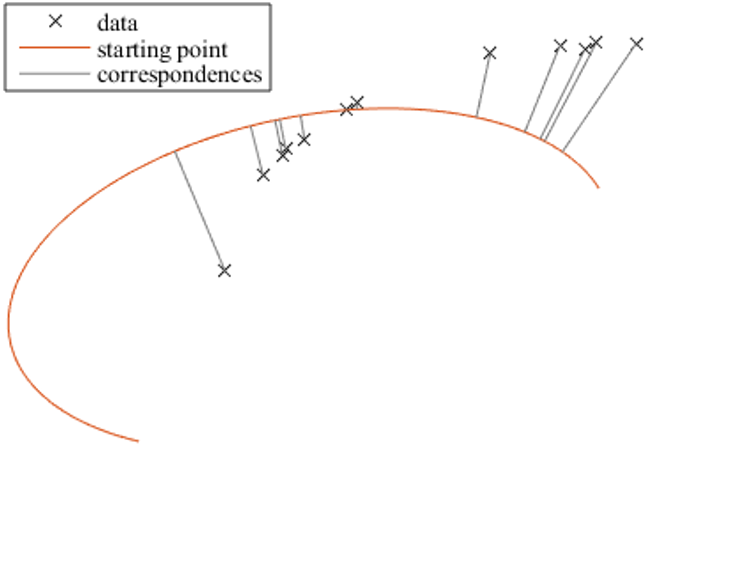}
		\caption{Initial point-matching.}
		\label{subfig:select}
	\end{subfigure}
	\begin{subfigure}[b]{0.5\linewidth}
	\centering
	\includegraphics[width=\linewidth]{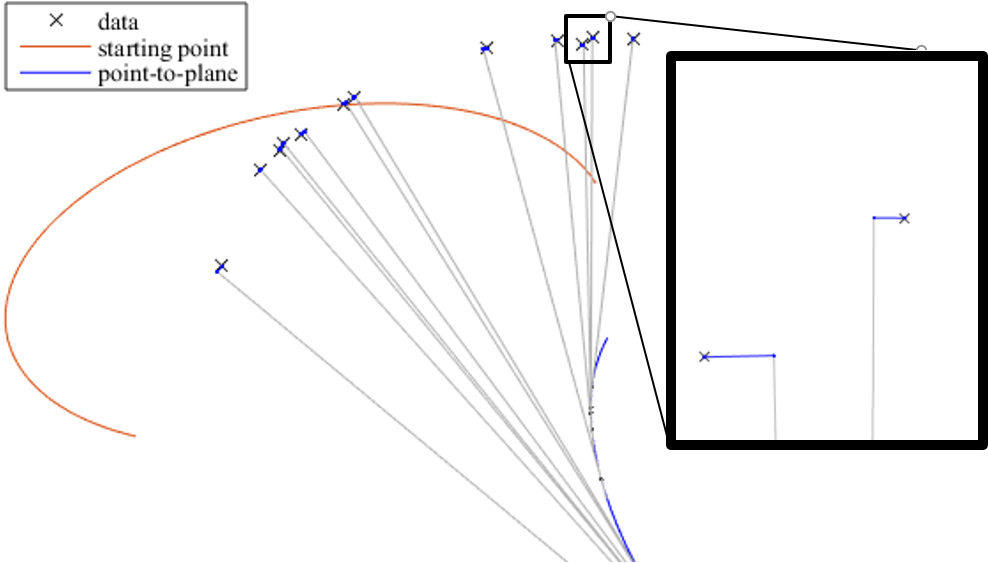}
	\caption{Single unconstrained Point-to-Plane ICP update.}
	\label{subfig:point-to-plane-icp}
	\end{subfigure}
	\begin{subfigure}[b]{0.5\linewidth}
		\centering
		\includegraphics[width=\linewidth]{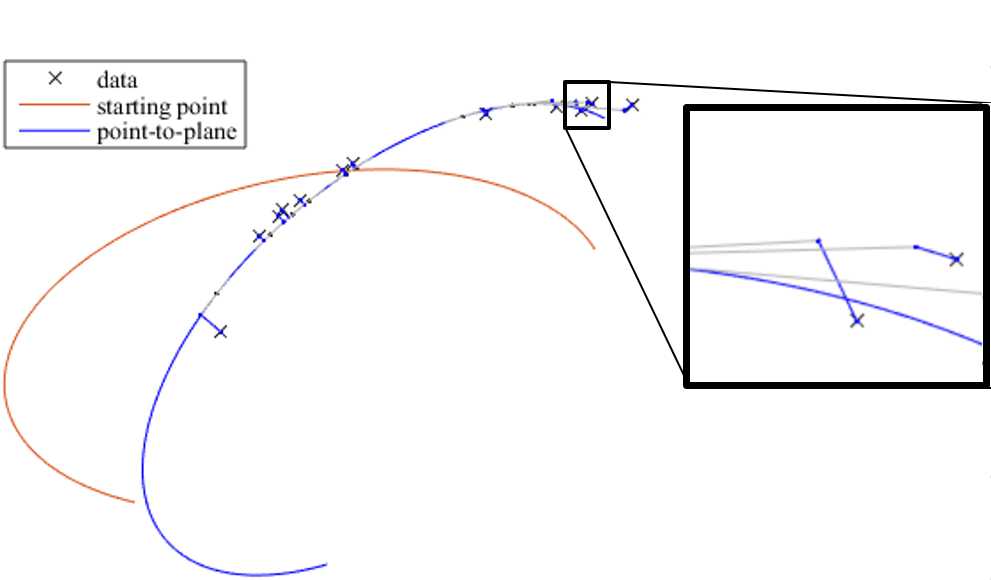}
		\caption{Single regularized Point-to-Plane ICP update.}
		\label{subfig:reg-point-to-plane-icp}
	\end{subfigure}
	\begin{subfigure}[b]{0.37\linewidth}
		\centering
		\includegraphics[width=\linewidth]{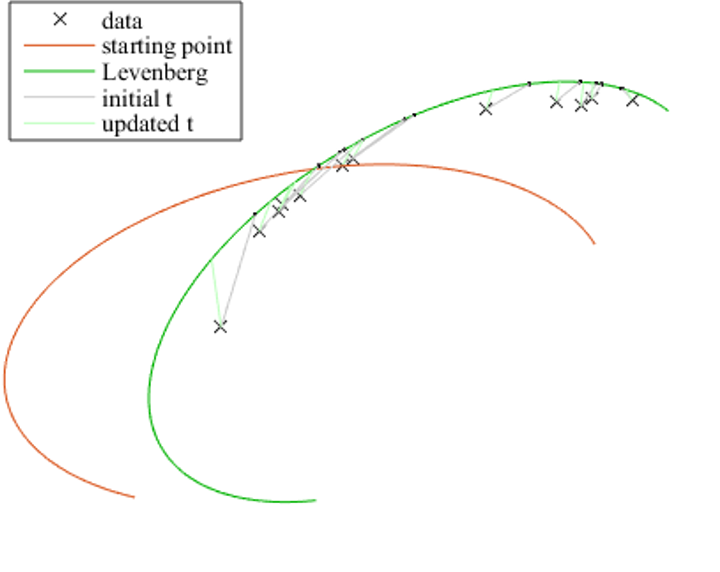}
		\caption{Single lifted update.}
		\label{subfig:lifted}
	\end{subfigure}
	\caption{Illustration of point-to-plane ICP and lifted optimization.}
	\label{fig:icp}
\end{figure}

We use the rigid alignment of a 2D curve $C$ to a set of data points $\{ {\mathbf x}_i \}^{N}_{i=1}$ to illustrate point-to-plane ICP updates, see Fig.~\ref{fig:icp}.
As point-to-plane ICP finds correspondences between the data and the tangent space of the model, in the 2D case we will be performing `point to tangent line' updates.
Let $\theta$ parametrize the translation and rotation of the curve, which are the parameters we want to solve for.

Both point-to-plane ICP and lifting first need to select the point-to-curve correspondences between $N$ data points and the posed curve $C(\theta)$. 
There are many point-matching variants~\cite{Rusinkiewicz2001}, and Fig.~\ref{subfig:select} depicts simple closest-point correspondences, but our analysis extends to any initial correspondence proposals.
Let us denote these correspondences by curve parameter positions $\{ {t}_i \}^{N}_{i=1}$, and thus data point ${\mathbf x}_i$ is paired with curve point $C(t_i,\theta)$.

The tangent line at model point $C(t_i)$ is $\{ C(t_i) + \gamma \dot{C}(t_i) | \gamma \in \mathbb{R} \}$, where $\dot{C}(t_i)$ denotes the unit-length tangent vector at $t_i$.
Then one completely unconstrained point-to-plane ICP update (Fig.~\ref{subfig:point-to-plane-icp}) finds $\theta^*$ that minimizes the distance of point-to-tangent-line:
\begin{equation}
\theta^* = \argmin_\theta \sum_{i} \min_{\gamma_i} {\| {\mathbf x}_i  - (C(t_i,\theta) + \gamma_i \dot{C}(t_i,\theta)) \|^2},
\label{eq:point-to-plane-icp}
\end{equation}
where $(C(t_i,\theta) + \gamma_i \dot{C}(t_i,\theta))$ is the projection of the data point on the tangent line when evaluated at the position $\gamma_i$ that minimizes the distance to ${\mathbf x}_i$, i.e., the footpoint.
Note that this is a 2D version of the original point-to-plane method by Chen and Medioni~\cite{Chen1991}, and it is equivalent to using the distance metrics based on the normal at $C(t_i,\theta)$ 
described by Low~\cite{Low04}.
As shown in Fig.~\ref{subfig:point-to-plane-icp}, this could lead to a bad update when the starting point is not close enough.

To do better, a regularized version of point-to-plane ICP (Fig.~\ref{subfig:reg-point-to-plane-icp}) could be to find $\theta^*$ that minimizes the distance of both point-to-point and point-to-tangent-line:
\begin{equation}
\theta^* = \argmin_\theta \sum_{i} \min_{\gamma_i} {\| {\mathbf x}_i  - (C(t_i,\theta) + \gamma_i \dot{C}(t_i,\theta)) \|^2} + \lambda {\| \gamma_i \|^2}.
\label{eq:reg-point-to-plane-icp}
\end{equation}
Here the inner minimization over $\{ {\gamma}_i \}^{N}_{i=1}$ can be solved separately as a series of least-squares problems, and $\theta$ solved afterwards in each iteration.

Finally, lifted optimization (Fig.~\ref{subfig:lifted}) simultaneously finds $\theta^*$ and $T = \{ {t}_i \}^{N}_{i=1}$ that minimizes
\begin{equation}
\theta^*, T^* = \argmin_{\theta,T} \sum_{i} {\| {\mathbf x}_i  - C(t_i,\theta) \|^2} 
\label{eq:lifted}
\end{equation}

We can see that point-to-plane ICP shares similarities with a lifted update; both allow a model to slide against the data in each iteration, improving convergence. 
While lifting achieves the sliding by simultaneously optimizing $\theta$ and $T$ in the update, point-to-plane ICP achieves the sliding by projection onto the model tangent plane.
However, point-to-plane ICP addresses this for only one energy formulation (point-to-model distances) whereas a lifted optimizer generalizes to arbitrary differentiable objectives.
Recall that we include a normal disparity term in our energy (Eq.\ 4 in main paper), which is critical for faster convergence as shown in the experiment described in Sec.~3.1 and Fig.~7 of the main paper;
point-to-plane ICP cannot minimize this objective.

\section{Efficiency of lifting vs. ICP}
\label{section:cmp-icp}


As shown in Fig.~6 (right) and Fig.~12 in main paper, our actual runtime for ICP is \emph{slower} than for lifting in terms of wall-clock timing on PC.
While we believe our implementations for both are reasonable, this efficiency comparison could still be unfair in many ways.

Instead, let us use FLOP (Floating Point Operation) counts to theoretically compare the two in the context of hand tracking.
Note that in the extreme low-power case we discuss (1\% of an iPhone 7), we assume the programmer is already optimizing cache hit rates etc.
In our code for hand tracking on HoloLens 2, for example, our tightest loop is running at 75\% of theoretical peak FLOPS.

Several of the steps of ICP are comparable to lifting:

\begin{enumerate}
\item A major component of the cost is determining the point-to-mesh correspondences between $N$ data points (less than $200$ for our hand tracking examples), and $M$ sampled mesh points.
For lifting, $M$ can be relatively small (e.g.\ 300), as most correspondences accept their lifted update.
We might hope that a spatial index over mesh samples would accelerate this per-data-point query, but as the mesh changes at every iteration, a KD-tree or other spatial data structure would need to be rebuilt at every iteration, and for $M\approx 300$ this cost is never recouped.
The cost is therefore $O(MN)$ FLOPs, although the constant factor can be small with efficient SIMD usage.


For ICP, as the correspondences are held fixed, convergence depends on finding good matches at every iteration.
The mesh samples therefore either need to be numerous, say $M=3000$, where a KD-tree can achieve asymptotic complexity $O(N \log M)$ but with a constant overhead that makes the incurred cost greater when amortized per query.
Or else ICP can use smaller $M$ but needs at least one Newton step per closest-point query, 
involving the solution of $N$ $2 \times 2$ linear systems adding to the $O(MN)$ cost above.

Parallelization is trivial for this step, but does not reduce the number of operations, and therefore cannot help to address the power constraints of a mobile device.
The impact on latency is important but it does not change the FLOP count analysis here.

\item The second major component is the cost of solving for the~$\theta$ update ($P$ params), and the $N$ correspondence updates in the case of lifting.
For ICP 
this involves a $P \times P$ linear system solve per sub-iteration at a cost of $O(P^3)$.
In lifting, the system is more complex but still quite sparse, adding $O(PN)$ operations.

Taylor et al.~\cite{Taylor2016} also point out in their Section 3.3.1 
that lifted optimization scales linearly with $N$; in fact,
the extra floating-point multiplications required for lifting to solve $\Delta \theta$ (their Eq.\ 22) and $\Delta U$ are about $(18+4P)N$, which with $N=128$ and $P=28$ as in our HoloLens hand tracking example gives only $\approx$ 16K additional floating-point multiplications per iteration ($<$2MFLOPS overall).
\end{enumerate}

So iteration timings are very comparable, and from our convergence figures, 
we emphasize that the second-order convergence of lifted optimization 
is much faster than the linear convergence observed for ICP.
For example, Fig.~6 (left) in the main paper shows that Lifted Phong/Subdiv (red/green solid line in (a)) converges within $\approx 13$ iterations, 
while the ICP Phong/Subdiv (red/green dashed line) within $\approx 50$ iterations, i.e.~{\bf 3.8 times more}.
For the hand tracking experiment in the main paper (Fig.~12(a)), to achieve an accuracy level where 79\% of dataset has average joint error $<$ 20mm,
Lifted Phong (red solid line) requires $\approx 4$ iterations, 
while ICP Phong (red dashed line) requires $\approx 7$ iterations, i.e.~{\bf 1.75 times more}.
Recall that from the analysis above, the per-iteration cost of lifting for $N \le 200$ is comparable to ICP.

In summary, depending on several factors ICP can range from marginally cheaper than lifted optimization per iteration to considerably more expensive, but also comes with the cost of increases in iteration count (and decrease in basin of convergence, requiring more compute for any initial estimate).

\section{More results on rigid pose estimation of an ellipsoid}

Here we show more results on convergence comparison of various surface types and optimization frameworks on rigid pose estimation of an ellipsoid (Sec~\ref{sec:toyEx} in main paper).

\noindent {\bf Qualitative comparisons.}
Fig.~\ref{fig:toyExVis} visualizes the optimization iterations to reach the ground-truth rigid pose $[0.1, 0.3, 2.0, 1, 1,1]$.
In the \emph{lifted} case, the \emph{Phong} and \emph{Subdiv} surfaces both converge to the correct pose within 5 iterations; \emph{Tri. mesh} needs 15 iterations.
With \emph{ICP}, the \emph{Phong} and \emph{Subdiv.} surfaces both converge to the correct pose within 24 iterations; \emph{Tri. mesh} needs 35 iterations.

\begin{figure*}[t]
	\input{figs/ToyEx/ellipsoidAccuracyBestF1280.tex}
	\caption{Rigid alignment accuracy for an ellipsoid with 1280 facets. The optimization runs for max. 50 iterations.}
	\label{fig:toyExAccuracyF1280}
\end{figure*}
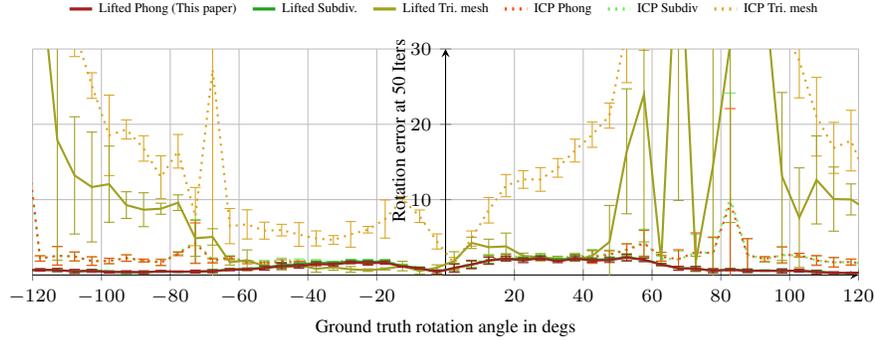

\begin{figure*}[t]
	\begin{center}
		\includegraphics[width=0.85\linewidth]{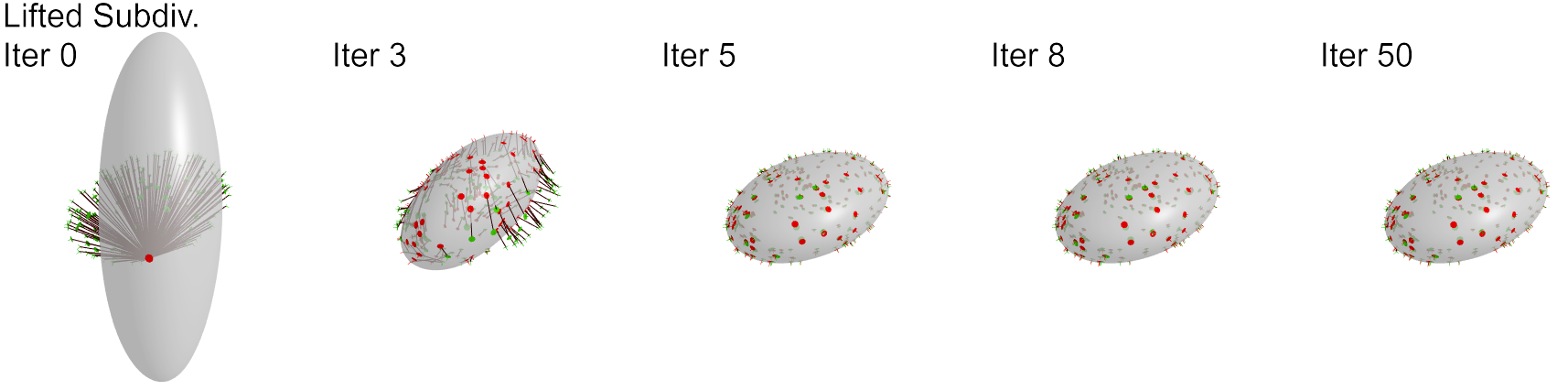}
		\includegraphics[width=0.85\linewidth]{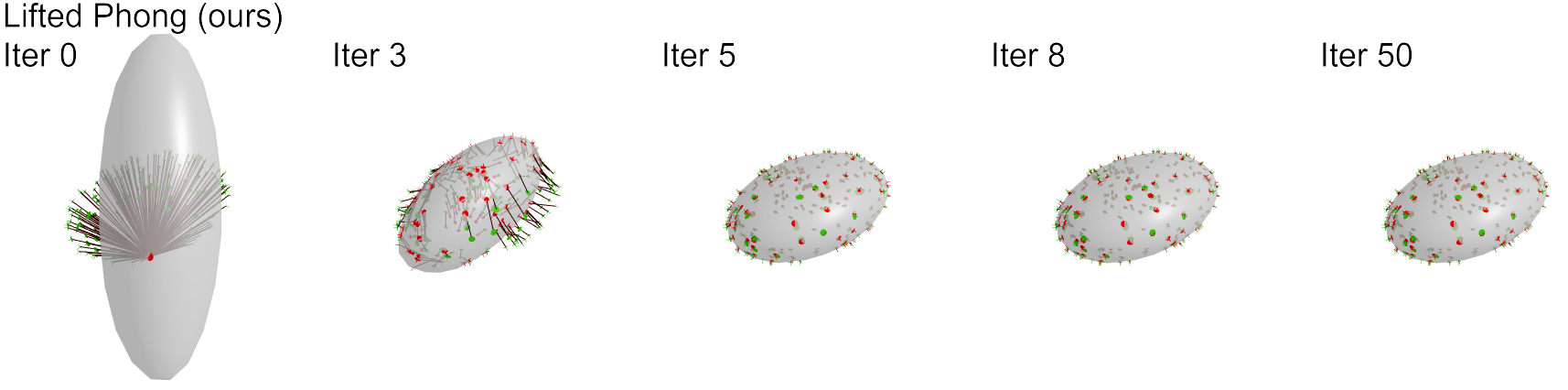}
		\includegraphics[width=0.85\linewidth]{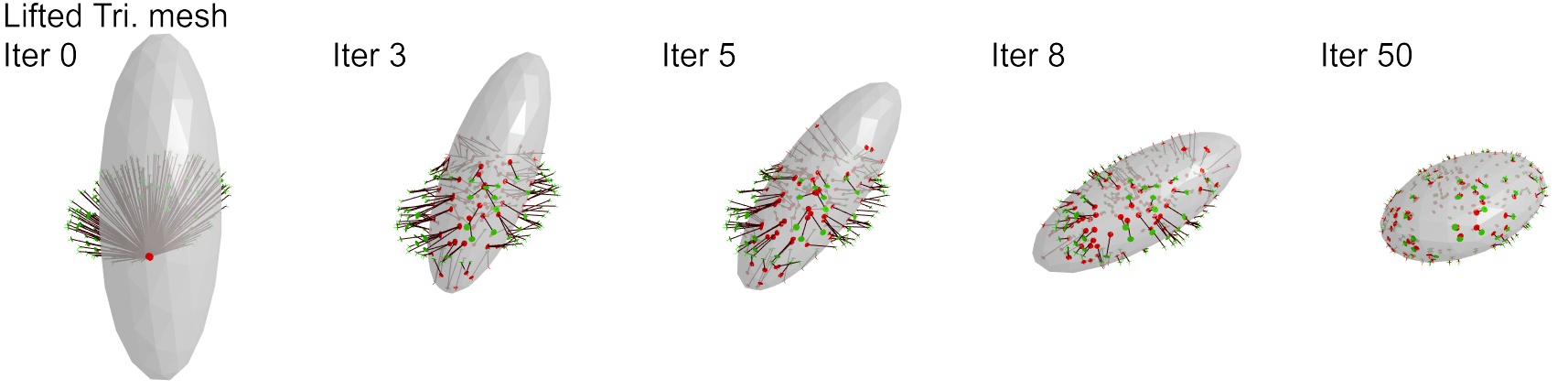}
		\includegraphics[width=0.85\linewidth]{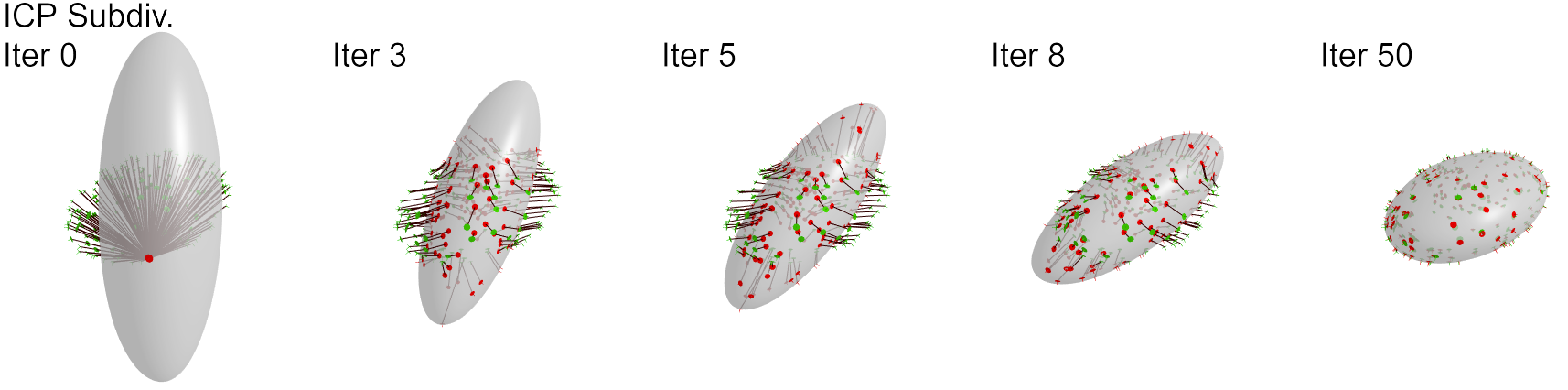}
		\includegraphics[width=0.85\linewidth]{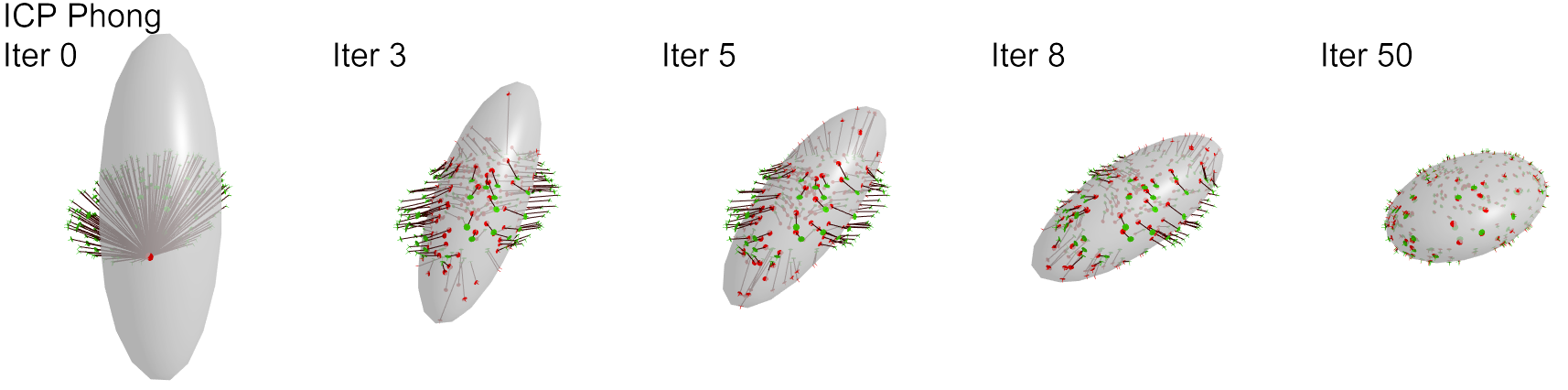}
		\includegraphics[width=0.85\linewidth]{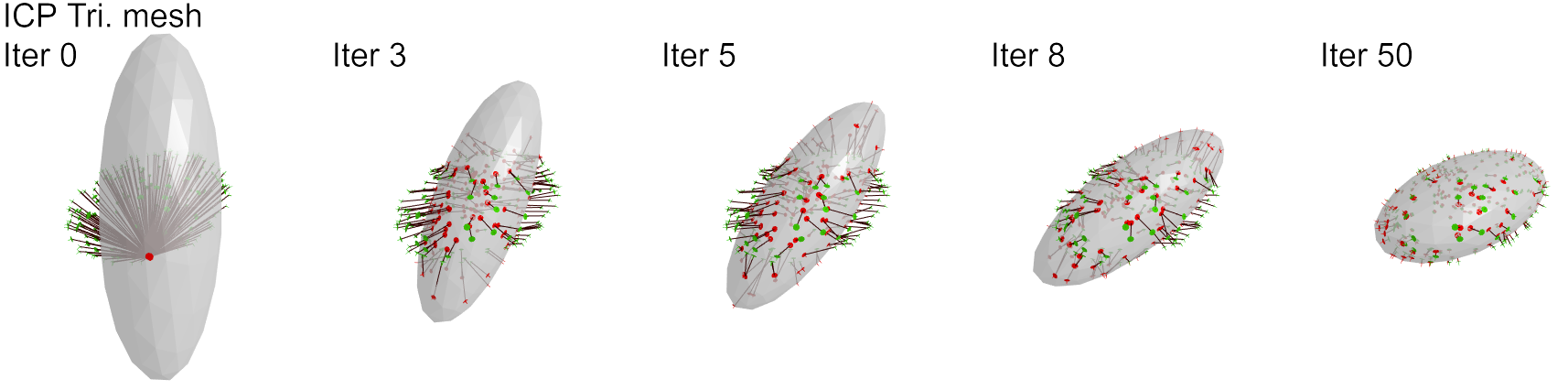}
	\end{center}
	\caption{ Rigid pose alignment of an ellipsoid with 3 surface types in lifted optimization and ICP optimization. The green and red dots represent data points and surface points, respectively; black lines denote correspondences between the two. Only the first two methods converge within 5 iterations.}
	\label{fig:toyExVis}
\end{figure*}

\noindent {\bf Varying mesh resolution.} 
As for the \emph{Phong} interpolated normals exhibit less variation with higher-resolution meshes, we run a similar experiment considering an ellipsoid with $4$x denser triangles.
Fig.~\ref{fig:toyExAccuracyF1280} shows that, in this case, \emph{Tri. mesh} get slightly better accuracy than with lower resolution. However, the overall performance seems comparable to that reported in Fig.~\ref{fig:toyExAccuracyF320}.

\noindent {\bf Noise in data normals.} 
As our proposed model-fitting method relies on data normal term for good convergence, one possible failure case will be when there is too much noise in the input data normals, due to either data too sparse or normal estimation too noisy. To confirm this, we tried to bump up the noise level in data normals in the ellipsoid example (Sec 3.1 and Fig.~5 in the main paper), from default random range [0.0, 0.1], up to [0.0, 0.25] and [0.0, 0.5]. We found that the \emph{Phong} does get slightly worse at 0.25, but still much better than \emph{Tri. mesh}; but the advantage gets much smaller at 0.5.

%% file: figs/ToyEx/ellipsoidAccuracyBestF1280.tex
%

\let \yaxiswidth \undefined
\newlength{\yaxiswidth}
\setlength{\yaxiswidth}{0.35\columnwidth}

\begin{minipage}[b]{\textwidth}
\begin{tikzpicture}
	\begin{axis}[
		xlabel={Ground truth rotation angle in degs}, 
		ylabel={Rotation error at 50 Iters},
		ylabel style={align=flush right,text width=\yaxiswidth,yshift=-0.15cm,xshift=-0.5cm},
		xticklabel={\pgfmathparse{1*\tick}$\pgfmathprintnumber{\pgfmathresult}$},
		yticklabel={\pgfmathparse{1*\tick}$\pgfmathprintnumber{\pgfmathresult}$},
		width=0.9\columnwidth,
		height=0.8\plotheight,
		ymin=0,
		ymax=30,
		xmin =-120,
		xmax=120,
		legend style={anchor=south,at={(0.5, 1.1)},/tikz/every even column/.append style={column sep=0.01\columnwidth}},
		legend columns=6,
		reverse legend
		]
\addplot+[chira3b, dotted, error bars/.cd, y fixed, y dir=both, y explicit, error bar style={solid}] table[x=x, y=y,y error=error] {figs/ToyEx/Data/ellipsoid_F1280/bestLambdas/ICP_PieceWiseLinear_err_2pi_iter50.data}; \addlegendentry{ICP Tri. mesh}
\addplot[chira2b, dotted, error bars/.cd, y fixed, y dir=both, y explicit, error bar style={solid}] table[x=x, y=y,y error=error] {figs/ToyEx/Data/ellipsoid_F1280/bestLambdas/ICP_ApproxLoopTriangleMonomialDeg4_err_2pi_iter50.data}; \addlegendentry{ICP Subdiv}
\addplot[chira1b, dotted, error bars/.cd, y fixed, y dir=both, y explicit, error bar style={solid}] table[x=x, y=y,y error=error] {figs/ToyEx/Data/ellipsoid_F1280/bestLambdas/ICP_PhongSurface_err_2pi_iter50.data}; \addlegendentry{ICP Phong}

\addplot[chira3, error bars/.cd, y fixed, y dir=both, y explicit, error bar style={solid}] table[x=x, y=y,y error=error] {figs/ToyEx/Data/ellipsoid_F1280/bestLambdas/Joint_PieceWiseLinear_err_2pi_iter50.data}; \addlegendentry{Lifted Tri. mesh}
\addplot[chira2, error bars/.cd, y fixed, y dir=both, y explicit, error bar style={solid}] table[x=x, y=y,y error=error] {figs/ToyEx/Data/ellipsoid_F1280/bestLambdas/Joint_ApproxLoopTriangleMonomialDeg4_err_2pi_iter50.data}; \addlegendentry{Lifted Subdiv.}
\addplot[chira1, error bars/.cd, y fixed, y dir=both, y explicit, error bar style={solid}] table[x=x, y=y,y error=error] {figs/ToyEx/Data/ellipsoid_F1280/bestLambdas/Joint_PhongSurface_err_2pi_iter50.data}; \addlegendentry{Lifted Phong (This paper)}
	
	\end{axis}
\end{tikzpicture}
\end{minipage}
\vskip-3pt

%% file: PhongSurface_eccv2020.bbl
\begin{thebibliography}{10}
\providecommand{\url}[1]{\texttt{#1}}
\providecommand{\urlprefix}{URL }
\providecommand{\doi}[1]{https://doi.org/#1}

\bibitem{Besl1992}
Besl, P.J., McKay, N.D.: A method for registration of 3-{D} shapes. IEEE
  Transactions on Pattern Analysis and Machine Intelligence  \textbf{14}(2),
  239--256 (Feb 1992). \doi{10.1109/34.121791},
  \url{http://dx.doi.org/10.1109/34.121791}

\bibitem{Bogo2016}
Bogo, F., Kanazawa, A., Lassner, C., Gehler, P., Romero, J., Black, M.J.: Keep
  It SMPL: Automatic Estimation of 3D Human Pose and Shape from a Single Image,
  pp. 561--578 (2016),
  \url{https://app.dimensions.ai/details/publication/pub.1052960414 and
  http://arxiv.org/pdf/1607.08128}

\bibitem{Cashman2013}
{Cashman}, T.J., {Fitzgibbon}, A.W.: What shape are dolphins? {B}uilding {3D}
  morphable models from {2D} images. IEEE Transactions on Pattern Analysis and
  Machine Intelligence  \textbf{35}(1),  232--244 (2013).
  \doi{10.1109/TPAMI.2012.68}

\bibitem{Chen1991}
{Chen}, Y., {Medioni}, G.: Object modeling by registration of multiple range
  images. In: Proceedings. 1991 IEEE International Conference on Robotics and
  Automation. pp. 2724--2729 vol.3 (April 1991).
  \doi{10.1109/ROBOT.1991.132043}

\bibitem{Chen1992}
Chen, Y., Medioni, G.: Object modelling by registration of multiple range
  images. Image and Vision Computing  \textbf{10}(3),  145--155 (1992)

\bibitem{fitzgibbon2003}
Fitzgibbon, A.: Robust registration of {2D} and {3D} point sets. In:
  Proceedings of the British Machine Vision Conference. pp. 411--420 (2001),
  \url{https://www.microsoft.com/en-us/research/publication/robust-registration-of-2d-and-3d-point-sets/}

\bibitem{Hirshberg2012}
Hirshberg, D., Loper, M., Rachlin, E., Black, M.: Coregistration: Simultaneous
  alignment and modeling of articulated {3D} shape. In: European Conf. on
  Computer Vision (ECCV). pp. 242--255. LNCS 7577, Part IV, Springer-Verlag
  (Oct 2012)

\bibitem{Khamis2015}
Khamis, S., Taylor, J., Shotton, J., Keskin, C., Izadi, S., Fitzgibbon, A.:
  Learning an efficient model of hand shape variation from depth images. In:
  Proc. CVPR. pp. 2540--2548 (2015)

\bibitem{Nikos2019}
Kolotouros, N., Pavlakos, G., Black, M., Daniilidis, K.: Learning to
  reconstruct {3D} human pose and shape via model-fitting in the loop. In: The
  IEEE Conference on Computer Vision and Pattern Recognition (CVPR) (09 2019)

\bibitem{Li2017}
Li, T., Bolkart, T., Black, M.J., Li, H., Romero, J.: Learning a model of
  facial shape and expression from {4D} scans. ACM Trans. Graph.
  \textbf{36}(6),  194:1--194:17 (Nov 2017). \doi{10.1145/3130800.3130813},
  \url{http://doi.acm.org/10.1145/3130800.3130813}

\bibitem{Loop1987}
Loop, C.T.: Smooth Subdivision Surfaces Based on Triangles. Master's thesis,
  University of Utah (August 1987),
  \url{http://research.microsoft.com/apps/pubs/default.aspx?id=68540}

\bibitem{Low04}
lim Low, K.: Linear least-squares optimization for point-toplane icp surface
  registration. Tech. rep. (2004)

\bibitem{MagicLeap2019}
{Magic Leap Inc}: {Perception at Magic Leap}.
  {https://sites.google.com/view/perceptionatmagicleap/} (June 2019)

\bibitem{marquardt1963}
Marquardt, D.W.: An algorithm for least-squares estimation of nonlinear
  parameters. Journal of the Society for Industrial and Applied Mathematics
  \textbf{11}(2),  431--441 (1963)

\bibitem{HoloLens2019}
{Microsoft}: {HoloLens 2}.
  {https://blogs.microsoft.com/blog/2019/02/24/microsoft-at-mwc-barcelona-introducing-microsoft-hololens-2}
  (2019)

\bibitem{Mueller2019}
Mueller, F., Davis, M., Bernard, F., Sotnychenko, O., Verschoor, M., Otaduy,
  M.A., Casas, D., Theobalt, C.: Real-time pose and shape reconstruction of two
  interacting hands with a single depth camera. ACM Trans. Graph.
  \textbf{38}(4),  49:1--49:13 (Jul 2019). \doi{10.1145/3306346.3322958},
  \url{http://doi.acm.org/10.1145/3306346.3322958}

\bibitem{neugebauer1997}
Neugebauer, P.J.: Geometrical cloning of {3D} objects via simultaneous
  registration of multiple range images. In: Proceedings of 1997 International
  Conference on Shape Modeling and Applications. pp. 130--139. IEEE (1997)

\bibitem{Pavlakos2019}
Pavlakos, G., Choutas, V., Ghorbani, N., Bolkart, T., Osman, A.A.A., Tzionas,
  D., Black, M.J.: Expressive body capture: {3D} hands, face, and body from a
  single image. In: Proceedings IEEE Conf. on Computer Vision and Pattern
  Recognition (CVPR) (2019)

\bibitem{Pellegrini2008}
Pellegrini, S., Schindler, K., Nardi, D.: A generalisation of the {ICP}
  algorithm for articulated bodies. In: Proceedings of the British Machine
  Vision Conference. pp. 87.1--87.10. BMVA Press (2008), doi:10.5244/C.22.87

\bibitem{Phong1975}
Phong, B.T.: Illumination for computer generated pictures. Commun. ACM
  \textbf{18}(6),  311--317 (Jun 1975). \doi{10.1145/360825.360839},
  \url{http://doi.acm.org/10.1145/360825.360839}

\bibitem{Qian2014}
Qian, C., Sun, X., Wei, Y., Tang, X., Sun, J.: Realtime and robust hand
  tracking from depth. pp. 1106--1113. IEEE (2014)

\bibitem{Rusinkiewicz2019}
Rusinkiewicz, S.: A symmetric objective function for {ICP}. ACM Trans. Graph.
  \textbf{38}(4),  85:1--85:7 (Jul 2019). \doi{10.1145/3306346.3323037},
  \url{http://doi.acm.org/10.1145/3306346.3323037}

\bibitem{Rusinkiewicz2001}
Rusinkiewicz, S., Levoy, M.: Efficient variants of the icp algorithm.
  Proceedings Third International Conference on 3-D Digital Imaging and
  Modeling pp. 145--152 (2001)

\bibitem{Sridhar2013}
Sridhar, S., Oulasvirta, A., Theobalt, C.: Interactive markerless articulated
  hand motion tracking using {RGB} and depth data. pp. 2456--2463 (Dec 2013),
  \url{http://handtracker.mpi-inf.mpg.de/projects/handtracker\_iccv2013/}

\bibitem{Sullivan1998}
Sullivan, S., Ponce, J.: Automatic model construction and pose estimation from
  photographs using triangular splines. IEEE Transactions on Pattern Analysis
  and Machine Intelligence  \textbf{20}(10),  1091--1097 (1998)

\bibitem{Tagliasacchi2015}
Tagliasacchi, A., Schr{\"o}der, M., Tkach, A., Bouaziz, S., Botsch, M., Pauly,
  M.: Robust articulated-{ICP} for real-time hand tracking  \textbf{34}(5),
  101--114 (2015)

\bibitem{Taylor2014}
Taylor, J., Stebbing, R., Ramakrishna, V., Keskin, C., Shotton, J., Izadi, S.,
  Hertzmann, A., Fitzgibbon, A.: User-specific hand modeling from monocular
  depth sequences. In: CVPR. pp. 644--651 (2014)

\bibitem{Taylor2016}
Taylor, J., Bordeaux, L., Cashman, T., Corish, B., Keskin, C., Sharp, T., Soto,
  E., Sweeney, D., Valentin, J., Luff, B., Topalian, A., Wood, E., Khamis, S.,
  Kohli, P., Izadi, S., Banks, R., Fitzgibbon, A., Shotton, J.: Efficient and
  precise interactive hand tracking through joint, continuous optimization of
  pose and correspondences. ACM Trans. Graph.  \textbf{35}(4),  143:1--143:12
  (jul 2016). \doi{10.1145/2897824.2925965},
  \url{http://doi.acm.org/10.1145/2897824.2925965}

\bibitem{Taylor2017}
Taylor, J., Tankovich, V., Tang, D., Keskin, C., Kim, D., Davidson, P., Kowdle,
  A., Izadi, S.: Articulated distance fields for ultra-fast tracking of hands
  interacting. ACM Trans. Graph.  \textbf{36}(6),  244:1--244:12 (Nov 2017).
  \doi{10.1145/3130800.3130853},
  \url{http://doi.acm.org/10.1145/3130800.3130853}

\bibitem{Tkach2016}
Tkach, A., Pauly, M., Tagliasacchi, A.: Sphere-meshes for real-time hand
  modeling and tracking. ACM Trans. Graph.  \textbf{35}(6),  222:1--222:11 (Nov
  2016). \doi{10.1145/2980179.2980226},
  \url{http://doi.acm.org/10.1145/2980179.2980226}

\bibitem{Wan2019}
Wan, C., Probst, T., Gool, L.V., Yao, A.: Self-supervised {3D} hand pose
  estimation through training by fitting. In: The IEEE Conference on Computer
  Vision and Pattern Recognition (CVPR) (June 2019)

\bibitem{Xiang2019}
Xiang, D., Joo, H., Sheikh, Y.: Monocular total capture: Posing face, body, and
  hands in the wild. CoRR  \textbf{abs/1812.01598} (2018),
  \url{http://arxiv.org/abs/1812.01598}

\bibitem{Zheng2014}
Zheng, J., Zeng, M., Cheng, X., Liu, X.: {SCAPE}-based human performance
  reconstruction. Computers \& Graphics  \textbf{38},  191--198 (2014)

\bibitem{ZuffiKB18}
Zuffi, S., Kanazawa, A., Black, M.J.: Lions and tigers and bears: Capturing
  non-rigid, 3d, articulated shape from images. In: 2018 {IEEE} Conference on
  Computer Vision and Pattern Recognition, {CVPR} 2018, Salt Lake City, UT,
  USA, June 18-22, 2018. pp. 3955--3963. {IEEE} Computer Society (2018).
  \doi{10.1109/CVPR.2018.00416},
  \url{http://openaccess.thecvf.com/content\_cvpr\_2018/html/Zuffi\_Lions\_and\_Tigers\_CVPR\_2018\_paper.html}

\bibitem{ZuffiKJB17}
Zuffi, S., Kanazawa, A., Jacobs, D.W., Black, M.J.: 3d menagerie: Modeling the
  3d shape and pose of animals. In: 2017 {IEEE} Conference on Computer Vision
  and Pattern Recognition, {CVPR} 2017, Honolulu, HI, USA, July 21-26, 2017.
  pp. 5524--5532. {IEEE} Computer Society (2017). \doi{10.1109/CVPR.2017.586},
  \url{https://doi.org/10.1109/CVPR.2017.586}

\end{thebibliography}
